\documentclass{article}

\usepackage{PRIMEarxiv}

\usepackage[utf8]{inputenc} % allow utf-8 input
\usepackage[T1]{fontenc}    % use 8-bit T1 fonts
\usepackage{hyperref}       % hyperlinks
\usepackage{url}            % simple URL typesetting
\usepackage{booktabs}       % professional-quality tables
\usepackage{amsfonts}       % blackboard math symbols
\usepackage{nicefrac}       % compact symbols for 1/2, etc.
\usepackage{microtype}      % microtypography
\usepackage{lipsum}
\usepackage{fancyhdr}       % header
\usepackage{graphicx}       % graphics
\graphicspath{{media/}}     % organize your images and other figures under media/ folder

\usepackage{amsmath}
\usepackage{color}
\usepackage{floatrow}
\usepackage{float}
\usepackage[dvipsnames]{xcolor}
\usepackage{pgfplots}
\pgfplotsset{height=6cm,compat=1.9}
\usepackage{epsfig}
\usepackage{amssymb}
\title{IAE-Net: Integral Autoencoders for  Discretization-Invariant Learning}

\author{
  Yong Zheng Ong, Zuowei Shen \\
  Department of Mathematics \\
  National University of Singapore \\
  21 Lower Kent Ridge Rd, Singapore 119077\\
  \texttt{\{e0011814, matzuows\}@u.nus.edu} \\
   \And
  Haizhao Yang \\
  Department of Mathematics\\
  University of Maryland College Park \\
  4176 Campus Dr., College Park, MD, 20742 \\
  \texttt{hzyang@umd.edu} \\
}

\begin{document}
\maketitle

\begin{abstract}
Discretization invariant learning aims at learning in the infinite-dimensional function spaces with the capacity to process heterogeneous discrete representations of functions as inputs and/or outputs of a learning model. This paper proposes a novel deep learning framework based on integral autoencoders (IAE-Net) for discretization invariant learning. The basic building block of IAE-Net consists of an encoder and a decoder as integral transforms with data-driven kernels, and a fully connected neural network between the encoder and decoder. This basic building block is applied in parallel in a wide multi-channel structure, which is repeatedly composed to form a deep and densely connected neural network with skip connections as IAE-Net. IAE-Net is trained with randomized data augmentation that generates training data with heterogeneous structures to facilitate the performance of discretization invariant learning. The proposed IAE-Net is tested with various applications in predictive data science, solving forward and inverse problems in scientific computing, and signal/image processing. Compared with alternatives in the literature, IAE-Net achieves state-of-the-art performance in existing applications and creates a wide range of new applications where existing methods fail.
\end{abstract}

% keywords can be removed

\section{Introduction}

Deep learning was originally proposed with neural networks (NN) parametrizing mappings from a vector space $X\subset \mathbb{R}^{d_X}$ to another vector space $Y\subset\mathbb{R}^{d_Y}$. In this learning setting, deep learning has become very successful in various applications not just limited to computer science. In many applications, the input space $X\subset \mathbb{R}^{d_X}$ and/or the output space $Y\subset \mathbb{R}^{d_Y}$ come from the discretization of infinite-dimensional function spaces. Therefore, NNs trained with fixed discrete spaces $X\subset \mathbb{R}^{d_X}$ and $Y\subset\mathbb{R}^{d_Y}$ cannot be applied to other discrete spaces associated with other discretization formats. For example, let $\mathcal{X}$ denote a Hilbert space of functions defined on $\Omega_x=[0,1]^{d_x}$ and let $\mathcal{Y}$ denote a Hilbert space of functions defined on $\Omega_y=[0,1]^{d_y}$. The goal is to learn a mapping $\Psi:\mathcal{X}\rightarrow \mathcal{Y}$. In the numerical implementation, for a fixed set of samples $\{x_1,\dots,x_{d_X}\}\subset [0,1]^{d_x}$, let $X=\{ \bar{u}\in\mathbb{R}^{d_X}:\bar{u}_i=u(x_i), u\in \mathcal{X}\}$; for a fixed set of samples $\{y_1,\dots,y_{d_Y}\}\subset [0,1]^{d_y}$, let $Y=\{ \bar{v}\in\mathbb{R}^{d_Y}:\bar{v}_i=v(y_i), v\in \mathcal{Y}\}$. Then learning $\Psi:\mathcal{X}\rightarrow \mathcal{Y}$ is reduced to learning $\bar{\Psi}:X\rightarrow Y$, where an NN $\Psi^n(\bar{u};\theta)$ with parameters $\theta$ is trained to approximate $\bar{\Psi}(\bar{u})$ for all $\bar{u}\in X$ in existing deep learning methods. In this sense, $\Psi^n(\bar{u};\theta)$ infers a good discrete representation of $\Psi(u)$. However, $\Psi^n$ cannot be applied to an arbitrary discrete representation of $u\in \mathcal{X}$, especially when $u$ has a discrete representation of dimension different to $d_X$. Similarly, the NN $\Psi^n$ cannot infer $v=\Psi(u)$ in an arbitrary discretization format. However, heterogeneous data structures are ubiquitous due to data sensing and collection constraints and, therefore, it is crucially important to design discretization-invariant learning. The goal of this paper is to show that:
\begin{itemize}
    \item Discretization invariance is achievable for mathematically well-posed problems even with simple interpolation techniques, but IAE-Net achieves the state-of-the-art accuracy.
    \item Existing methods fail to achieve discretization invariance for ill-posed and highly oscillatory problems, while IAE-Net succeed with reasonablly good accuracy. 
\end{itemize}

There are several immediate solutions to deal with heterogeneous data structures. Typically, expensive re-training is conducted to train a new NN for different discretization formats. However, re-training sometimes would be too expensive for large-scale NNs, or training samples from a certain discrete format might be too limited to obtain good accuracy. Another simple idea is to pre-process the input data and post-process the output data, e.g., using resizing, padding, interpolation \cite{bicubicinterpolation}, or cropping \cite{precipitation} to make data fit the specific requirement of an NN model. However, the numerical accuracy of these simple methods is not satisfactory in challenging applications as we shall see in the numerical experiments. Resizing signals/images to a specific size without deforming patterns contained therein is a major challenge. There are two cases in resizing: downsampling and upscaling. Downsampling may significantly reduce the vital features of data and make it more challenging to learn. Upscaling may generate side effects that mislead NNs. %rnn and lstm?

The above drawbacks have motivated the development of advanced frameworks for discretization-invariant learning recently. There are two main advanced ideas to achieve discretization-invariant transforms in deep learning. The first one is to apply ``pointwise" linear transforms in deep neural networks. Traditionally, a linear transform in NNs is a matrix-vector multiplication with a fixed matrix size, and, hence, the linear transform cannot be applied to input vectors with different dimensions, resulting in discretization dependence. The pointwise linear transform to a vector $\bar{u}\in\mathbb{R}^{d_X}$ is to apply the same linear transform to each entry of $\bar{u}$ so that the numerical computation can still be implemented for different dimension $d_X$'s. This technique is widely used in natural language processing, e.g., the transformer \cite{attention} and its variants \cite{bert,afno,leethorp2021fnet,liu2021swin,metaformer}, recurrent architectures \cite{rnn2,lstm,lstmforget,rnn,seq2seq}, and is recently applied to scientific computing for solving parametric partial differential equations (PDEs) and initial value problems \cite{fno,galerkintransformer,gupta2022nonlinear}. 

Another approach is to apply a parametrized linear integral transformation in NNs, e.g., $v(y)=\int_{\Omega_x}K(x,y;\theta)u(x)dx$ as a linear transformation from $u$ to $v$ in each layer of a deep NN, where $K(x,y;\theta)$ is an integral kernel parametrized with $\theta$. The integral can be implemented independent of the discretization of $u(x)$ so that the integral transform can be applied to arbitrary discrete representation $\bar{u}\in\mathbb{R}^{d_X}$ of $u(x)\in\mathcal{X}$, i.e., $d_X$ is allowed to be different. Purely convolutional NNs \cite{unet,guocnn,ellipses,bayesian,bhatnagar,khoo2021solving} without any fully connected layers fall into this approach, where the integral kernel is a convolutional kernel parametrized by a few parameters. NNs have also been applied to parametrize $K(x,y;\theta)$ for image classification and learning the mathematical operators to solve parametric PDEs and initial value problems \cite{chenchen,deeponet,deeponet2,li2020neural,fno,li3neural,li4neural,patel2021,you2022nonlocal}.

There have been mainly two research directions to theoretically investigate discretization invariant learning, especially for learning an operator from an infinite-dimensional space to another. In terms of approximation theory for operators, based on the universal approximation theory of operators \cite{chenchen} and the quantitative deep network approximation theory \cite{yarotsky2017error,yarotsky18a,shijun2,shijun3,shijun6}, there have been several works explaining the power of deep neural networks for approximating operators quantitatively \cite{bhattacharya2021model,kovachki2021universal,lanthaler2022error,mhaskar2022local,liu2022deep}. Especially in certain scientific computing problems, it was shown by \cite{deng2021convergence} that deep network approximation for solution operators mapping low-dimensional functions to another has no curse of dimensionality, even if the function spaces are infinite-dimensional. Going one step further to the generalization error analysis of neural network based operator learning, leveraging recent theories \cite{anthony1999neural,bartlett2019nearly,DBLP:journals/corr/abs-1809-03062,shin2020convergence,Luo2020,mishra2020estimates,lu2021priori,duan2021convergence,gu2021stationary}, there has been a posteriori and a priori generalization error analysis for learning nonlinear operators with discretization invariance by \cite{lanthaler2022error} and \cite{liu2022deep}, respectively.

Although discretization-invariant learning has been explored in the seminal works summarized above, these methods seem to focus on mathematically well-posed problems with non-oscillatory data, and cannot be applied to either ill-posed learning problems or problems with highly oscillatory data. This motivates the design of IAE-Net as a universal discretization-invariant learning framework for both well-posed and ill-posed problems with various data patterns. Besides the state-of-the-art accuracy in existing applications, IAE-Net has broad applications in science and engineering that existing methods are not capable of. For example, other than solving parametric PDEs \cite{deeponet,fno,wang2021learning,khoo2021solving}, in inverse scattering problems \cite{switchnet,wei2019physics}, it is interesting to learn an operator mapping the observed data function space to the parametric function space that models the underlying PDE for high-frequency phenomena. %In electron structure calculation, it is a fundamental problem to learn a nonlinear operator mapping a potential function to a density function \cite{MNN}. 
In imaging science \cite{phase}, an imaging process is to construct a nonlinear operator from the observed data function space to the reconstructed image function space. In image processing problems, e.g., image super-resolution \cite{resolution}, image denoising \cite{ellipses,Tian_2020}, image inpainting \cite{QIN2021102028}, the interest is to learn operators from an image discretized from one function to another. In signal processing, it is a fundamental problem to learn a nonlinear operator mapping a signal function to another or several signal functions \cite{fecgsyndb,scss,ecgseparation,waveunet,cass,kadandale2020multichannel,YANG202125}. 

In terms of technical innovation, there are four novel ideas in the design and training of IAE-Net to achieve state-of-the-art accuracy and broaden the application of discretization invariant learning to challenging problems. The main innovation can be summarized below.

\begin{figure}[t]
    \centering
    \includegraphics[width=0.8\columnwidth]{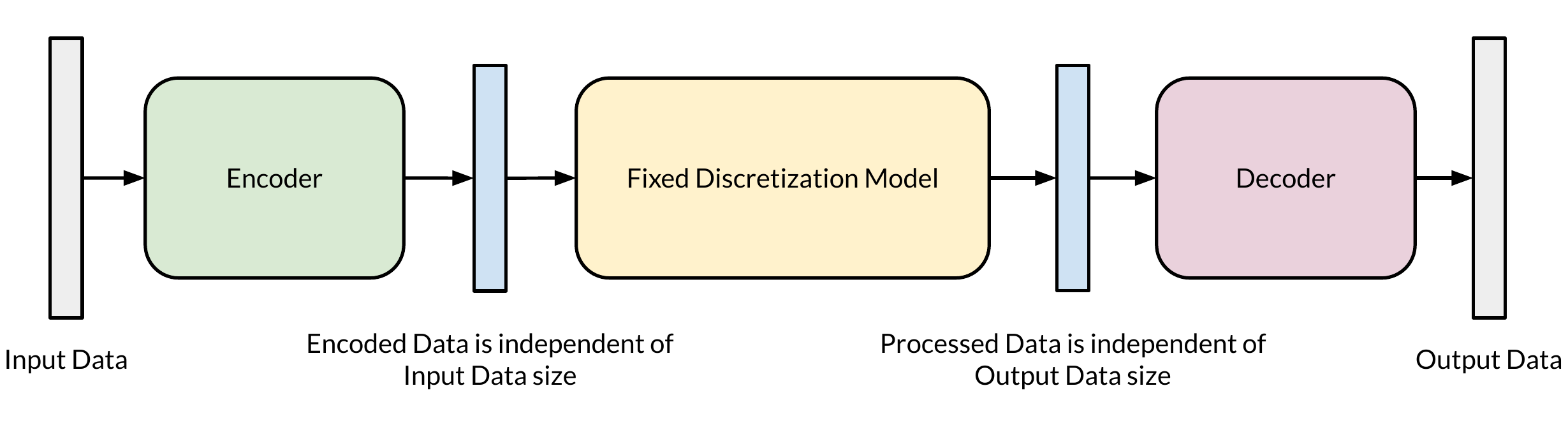}
    \caption{The visualization of a discretization invariant integral autoencoder structure. The design will be described in Section \ref{sec:iae}.}
    \label{fig:modelflow}
\end{figure}

\begin{figure}[t]
    \centering
    \includegraphics[width=0.8\columnwidth]{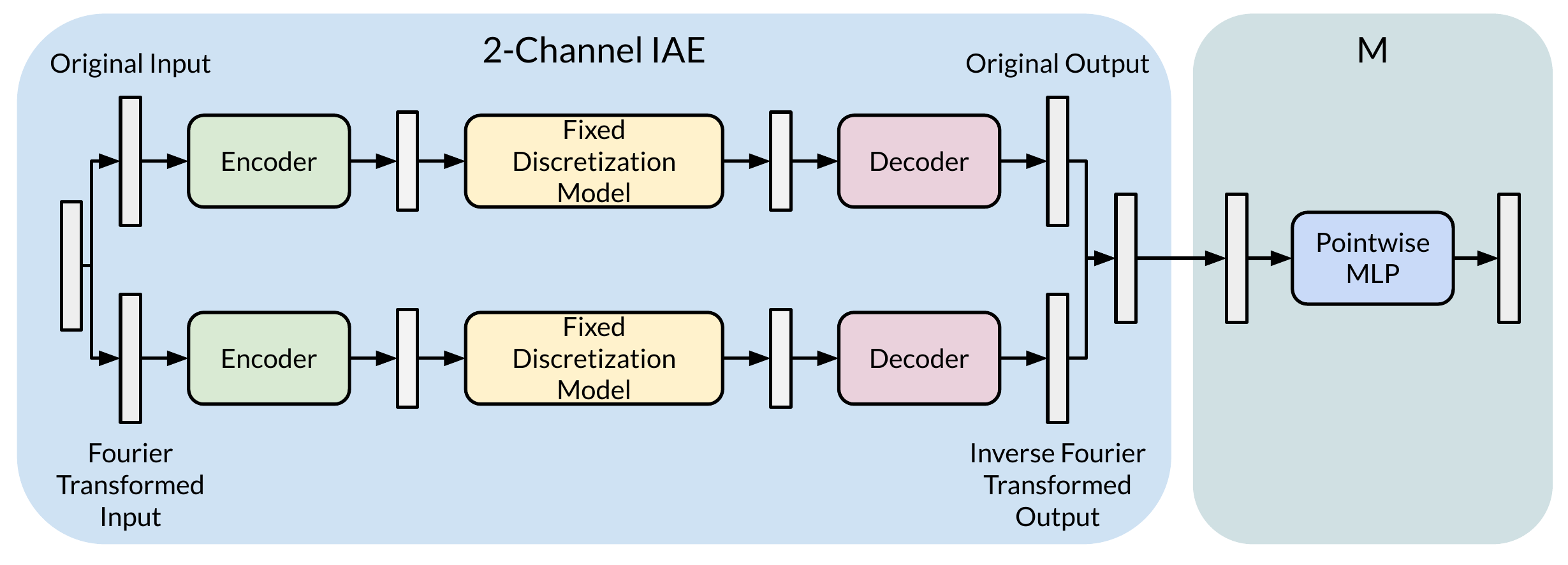}
    \caption{A $2$-channel IAE block with a pointwise MLP for merging parallel channels to implement Equation \eqref{eqn:multichannel}. The design will be described in Section \ref{sec:mcl}.}
    \label{fig:parallel}
\end{figure}

\begin{enumerate}
    \item For the first time, we introduce a data-driven integral autoencoder (IAE), as visualized in Figure \ref{fig:modelflow}, to achieve two important goals simultaneously: 1) IAE is discretization invariant via novel nonlinear integral transforms as the encoder and decoder; 2) IAE can capture the low-dimensional or low-complexity structure of a learning task for dimension reduction in the same spirit of data-driven principle component analysis (PCA). The integral kernels in these integral transforms are parametrized by NNs and trained to map functions in an infinite-dimensional space to a low and finite dimensional vector space (or vice versa). The nonlinearity, as oppose to linear integral transforms in existing works, is a key to improve performance. Due to the powerful expressiveness of NNs, these integral transforms could be trained to mimic useful transforms in mathematical analysis, e.g., the Fourier transform \cite{spectral}, wavelet transforms \cite{wavelet,daubechies1992ten,waveletbook,waveletbook2}, nonlinear PCA \cite{Scholz2002Nichtlineare,10.1007/978-3-540-73750-6_2}, etc.
    \item To make IAE-Net capable of solving problems with oscillatory phenomena, IAEs are applied in parallel, forming a multi-channel IAE block as shown in Figure \ref{fig:parallel}. Each channel has an IAE applied to data in a certain transformed domain. For example, Figure \ref{fig:parallel} shows an example of two IAEs, one of which is applied to map functions in their original domain, and the other one of which is applied to map functions in the frequency domain after Fourier transform. This multi-channel IAE structure can be extended to have AEs in different domains inspired by problem-dependent physical models.
    \item Inspired by DenseNet \cite{densenet}, by composing multi-channel IAE blocks with pointwise linear transforms as skip connections, a discretization-invariant densely connected structure is proposed as our final IAE-Net visualized in Figure \ref{fig:comparison_structure}. IAE-Net enjoys similar properties of the original DenseNet \cite{densenet} to boost the learning performance: lessening the gradient vanishing issue, diversifying the features of the inputs of each IAE block, and maximizing the gradient flow information of each IAE block directly to the final loss function. 
    \item IAE-Net is trained with randomized data augmentation that generates training data with heterogeneous structures to facilitate the performance of discretization invariant learning. In each iteration of the stochastic gradient descent (SGD) for training IAE-Net, the selected training samples are augmented with a set of samples randomly extrapolated/interpolated from the selected samples. This new data augmentation can be understood as a randomized loss function with a random extrapolation/interpolation operator with an image consisting of all possible discrete representations of the training data.
\end{enumerate}

In terms of scientific applications, IAE-Net can be applied not only to traditional machine learning problems, e.g., signal and image processing, classification, prediction, etc, but also to newly emerged scientific machine learning problems, e.g., solving PDEs, solving inverse problems, etc. In the literature of discretization invariant learning, existing algorithms \cite{fno, galerkintransformer} can only be applied to solve well-posed problems, e.g., initial value problems and parametric PDEs, aiming at a small zero-shot generalization error, i.e., NNs trained with data discretized from a fixed format can also be applied to other data with different discretization formats. Table \ref{tab:summary} compares the properties and performance of IAE-Net against the baseline models used in this paper. In this paper, we explore the application of IAE-Net to one more situation: the training data are generated with different discretization formats and each format is only applied to generate a small set of training data. This new setting is to simulate the computational environment of many data science problems with heterogeneous training data due to data sensing and collection constraints. Learning in this new setting can also alleviate data cleaning difficulty and cost in data science.

\begin{table}[H]
    \centering
    \caption{Comparison of different DNNs. ResNet: \cite{resnet}. DeepOnet: \cite{chenchen,deeponet}. Fourier Neural Operator (FNO): \cite{fno}. Fourier Transformer (FT), Galerkin Transformer (GT): \cite{galerkintransformer}. IAE-Net: designed in this proposal. "Yes" means $L2$ relative error (Equation \ref{eqn:l2loss}) is lower than $10\%$ for the trained resolution; otherwise, we mark "No". DeepONet is labelled $^\dag$ as it is discretization invariant and can handle irregular grids only for the output, but not the input.}
    \begin{tabular}{c|ccccccc}
        \textbf{Model} & \textbf{Discretization} & \textbf{Irregular} & \textbf{Predic-} & \textbf{Solve} & \textbf{Inverse} & \textbf{Image} & \textbf{Signal} \\
        \textbf{} & \textbf{Invariance} & \textbf{Grids} & \textbf{tion} & \textbf{PDE} & \textbf{Problem} & \textbf{Process} & \textbf{Process} \\
        \hline
        ResNet & No & No & \textbf{Yes} & \textbf{Yes} & \textbf{Yes} & \textbf{Yes} & No \\
        DeepONet$^{\dag}$ & No & No & \textbf{Yes} & \textbf{Yes} & No & No & No \\
        FNO & \textbf{Yes} & No & \textbf{Yes} & \textbf{Yes} & No & No & No \\
        FT & \textbf{Yes} & No & \textbf{Yes} & \textbf{Yes} & No & No & No \\
        GT & \textbf{Yes}& No & \textbf{Yes} & \textbf{Yes} & No & No & No \\
        \hline
        \textbf{IAE-Net} & \textbf{Yes} & \textbf{Yes} & \textbf{Yes} & \textbf{Yes} & \textbf{Yes} & \textbf{Yes} & \textbf{Yes}\\
    \end{tabular}
    \label{tab:summary}
\end{table}

In terms of computational methodologies, IAE-Net can be considered as an algorithm compression technique that can transform the research goal of computational science in many applications. For example, in scientific computing, solving inverse problems accurately and efficiently is still an active research field. Once trained, IAE-Net can solve inverse problems rapidly with a few matrix-vector multiplications. Therefore, it may be sufficient to focus on designing accurate algorithms for inverse problems without much attention to the computational efficiency. These accurate algorithms can generate training data for IAE-Net, and the trained IAE-Net can predict output solutions of these algorithms efficiently, in which sense IAE-Net is a compressed and accelerated version of a slow but accurate algorithm. 

We have focused on discretization invariant learning from one function space to another. It is easy to generalize IAE-Net to learn a mapping from a function space to a finite dimensional vector space, or a mapping from a finite dimensional vector space to a function space, by composing IAE-NET with a fixed-size NN. The rest of this paper is organized as follows. IAE-Net and its training algorithm will be introduced in detail in Section \ref{sec:alg}. Section \ref{sec:experiments} presents the experimental design and results for the applications covered in this paper. We conclude this paper with a short discussion in Section \ref{sec:con}.

\section{Algorithm Description}\label{sec:alg}

This section presents the detailed description of the design of IAE-Net.

\subsection{Problem Statement}
\label{sec:problemsetup}

IAE-Net targets general applications involving the task of learning an objective mapping $\Psi:\mathcal{X}\rightarrow \mathcal{Y}$, where $\mathcal{X}$ and $\mathcal{Y}$ are appropriate function spaces defined on $\Omega_x=[0,1]^{d_x}$ and $\Omega_y=[0,1]^{d_y}$ respectively. Here, $d_x$ and $d_y$ represent the number of dimensions of functions in each of the function spaces $\mathcal{X}$ and $\mathcal{Y}$ respectively. For example, for solving parametric PDEs, $\mathcal{X}$ refers to the coefficient function space while $\mathcal{Y}$ represents the solution function space. In signal separation, $\mathcal{X}$ refers to the space of 1D mixture signals while $\mathcal{Y}$ refers to the space of different extracted signals in the mixture. We denote $f\in \mathcal{X}$ and $g\in \mathcal{Y}$ as functions from each of the function spaces. The objective of discretization invariant learning is to learn the underlying mapping $\Psi$ using an NN $\Psi^{n}(\cdot;\theta):\mathcal{X}\rightarrow \mathcal{Y}$, where the NN parameters $\theta$ will be identified via supervised learning. In this paper, we propose IAE-Net as our new model for $\Psi^{n}(\cdot;\theta)$.

To work with the functions $f$ and $g$ numerically, we assume access only to the point-wise evaluations of the functions. This assumption is typical in real applications, where only access to a finite representation of the function is observable. It is simple to generalize the proposed discretization invariant learning to other discretization formats. The only difference is how the numerical evaluation of integral transforms in IAE-Net adapts to different discretization formats. Let $S_x=\{x_j\}_{j=1}^{s_x}\subset[0,1]^{d_x}$ and $S_y=\{y_j\}_{j=1}^{s_y}\subset[0,1]^{d_y}$ be two sets of sampling grid points containing $s_x$ and $s_y$ sampling points in $\Omega_x$ and $\Omega_y$, respectively. A numerical observation $\bar{f}$ of $f$ based on $S_x$ is thus defined as $\bar{f}=(f(x_1),\dots,f(x_{s_x}))$. Similarly, $\bar{g}=(g(y_1),\dots,g(y_{s_y}))$ defines a numerical observation of $g$ based on $S_y$. Thus, training data pairs are obtained as $p_{data}:=\{(\bar{f}_i,\bar{g}_i)\}_{i=1}^n$, where $n$ denotes the number of training data pairs. In our discretization invariant learning, $s_x$ and $s_y$ are allowed to be different for different observations $\bar{f}_i$ and $\bar{g}_j$ for any $i$ and $j$. Hence, $\Psi^{n}(\cdot;\theta)$ is not restricted to only a fixed discretization format. 

\subsection{Preliminary Definitions}
\label{sec:definitions}

This section introduces several definitions repeatedly used in this paper. 

\subsubsection{Fully-connected Neural Networks (FNNs)}

One kind of popular NNs is the fully-connected NN (FNN). An $N$-layer (or $N-1$-hidden layer) FNN is defined as the composition of $N$ consecutive affine transforms $A_{n_i,n_{i+1}}:\mathbb{R}^{n_i}\rightarrow\mathbb{R}^{n_{i+1}}$, $i=0,\dots,N-1$, and $N$ nonlinear activation functions. Here, given $x\in\mathbb{R}^{n_{i+1}}$, $A_{n_i,n_{i+1}}(x;W,b)=Wx+b$ for some learnable weight matrix $W\in\mathbb{R}^{n_{i+1}\times n_{i}}$ and bias $b\in\mathbb{R}^{n_{i+1}}$. More precisely, compositions are applied via the equation below
\begin{equation}\label{eqn:fnn}
    \Psi(x;\theta):=\sigma(A_{n_{N-1},n_{N}}(\sigma(A_{n_{N-2},n_{N-1}}(\dots(\sigma(A_{n_0,n_1}(x))))))),
\end{equation}
where $x\in\mathbb{R}^{n_0}$, $\sigma$ is an activation function, and $\theta$ consists of all parameters in the weight matrices and bias vectors. Since the design of $W$ in $A_{n_0,n_1}$ depends on $n_0$, the dimension of $x$, the FNN is non-discretization invariant. In this paper, if an FNN is used in IAE-Net, we will explain why the use of FNNs does not violate the discretization invariant property.

\subsubsection{Pointwise Multilayer Perceptrons (MLPs)}

Another commonly used building block of NNs are pointwise (or channel) multilayer perceptrons (MLPs). Suppose the input $x$ of an MLP has two dimenions: one is called the channel dimension, and the other one is called the spatial dimension that may have different sizes for different $x$'s. If $x$ has more than two dimensions, fixing the spatial dimension, all remaining dimensions can be reshaped into one channel dimension. For simplicity, we focus on the case when $x$ has two dimensions. An $N$-layer (or $N-1$-hidden layer) pointwise MLP is defined as the composition of $N$ consecutive affine transforms $B_{n_i,n_{i+1}}:\mathbb{R}^{n_i\times s}\rightarrow\mathbb{R}^{n_{i+1}\times s}$, $i=0,\dots,N-1$, and $N$ nonlinear activation functions. Here, given $x\in\mathbb{R}^{n_{i+1}\times s}$, $B_{n_i,n_{i+1}}(x;W,b)=Wx+b$ for some learnable weight matrix $W\in\mathbb{R}^{n_{i+1}\times n_{i}}$ and bias $b\in\mathbb{R}^{n_{i+1}}$. The compositions in the pointwise MLP are applied via the equation below
\begin{equation}\label{eqn:mlp}
   \Psi(x;\theta):= \sigma(B_{n_{N-1},n_{N}}(\sigma(B_{n_{N-2},n_{N-1}}(\dots(\sigma(B_{n_0,n_1}(x))))))),
\end{equation}
where $x\in\mathbb{R}^{n_0\times s}$, $\sigma$ is an activation function, and $\theta$ consists of all the weight matrices and bias vectors. Here, $n_0$ refers to the number of channels of $x$, and $s$ is the dimension of the input $x$ depending on discretization. For example, a colored image can be viewed as a $3\times s$ matrix where each channel represents the red, green and blue values, and $s$ represents the resolution of this image that may vary among different images. For each $i$, $i=1,\dots,s$, the pointwise MLP performs the same nonlinear transform to the $i$-th column of the input $x$. Hence, the pointwise MLP does not depend on the spatial resolution of $x$ and is thus naturally discretization invariant. This is in contrast with Equation \eqref{eqn:fnn}, which depends on the discretization of the input. A \textbf{pointwise linear transform} is defined as a $1$-layer pointwise MLP without $\sigma$.

\subsubsection{Integral Transform}
\label{sec:integral}

A key component of IAE-Net is the design of data-driven integral transforms. An integral transform of a function $f$ defined on $\Omega_x$ is defined by the transform $T$ of the form
\begin{equation}\label{eqn:integraltransform}
    (Tf)(z)=\int_{\Omega_x}K(x,z)f(x)dx,\quad z\in \Omega_z,
\end{equation}
where $K$, defined on $\Omega_x\times\Omega_z$ with $\times$ defining a cartesian product, is the kernel function of the transform. $Tf$ is a function on $\Omega_z$. 
%Similarly, the backward integral transform for a function $u$ defined on $\Omega_z$ is designed using another transform $T^{-1}$, via the following equation
%\begin{equation}\label{eqn:inverseintegraltransform}
%    (T^{-1}f)(x)=\int_{\Omega_z}K_2(x,z)u(z)dz,\quad x\in \Omega_x,
%\end{equation}
%where $K_2$ is another kernel function also defined on $\Omega_x\times\Omega_z$, and $T^{-1}f$ is a function of $\Omega_x$. $\Omega_z$ can be chosen to be independent of $\Omega$, and the forward and backward integral transform provides an avenue to transform the original input function to this independent domain and back, thus making it an excellent choice to design discretization invariant networks.

Equations \eqref{eqn:integraltransform} can be implemented numerically using a matrix-vector multiplication. Instead of the original function $f$, we are given observations $\bar{f}$ sampled from $f$ on $s$ grid points $S_x=\{x_i\}_{i=1}^s$ on $\Omega_x$. Similarly, let $S_z=\{z_j\}_{j=1}^m$ be $m$ grid points on $\Omega_z$. Thus, Equation \eqref{eqn:integraltransform} is implemented through a summation
\begin{equation}\label{eqn:numintegraltransform}
    (Tf)(z_j)=\frac{1}{s}\sum_{i=1}^sK(x_i,z_j)f(x_i),\quad z_j\in S_z,
\end{equation}
which is implemented via a matrix-vector multiplication. In particular, the computation of $Tf(z)$ for $z=z_1,\dots,z_m$ can be vectorized by assembling a kernel matrix $K$ with $K(x_j, z_i)$ as the $(i,j)$-th entry of $K$, and then the following matrix-vector multiplication is performed to compute $Tf$:
\begin{equation}\label{eqn:dotproductoriginal}
    \begin{pmatrix}
        (Tf)(z_1) \\ \dots \\ (Tf)(z_m)
    \end{pmatrix}=K\begin{pmatrix}
        f(x_1) \\ \dots \\ f(x_s)
    \end{pmatrix}=
    \begin{pmatrix}
        K(x_1, z_1) & \dots & K(x_s, z_1) \\
        & \dots & \\
        K(x_1, z_m) & \dots & K(x_s, z_m)
    \end{pmatrix}
    \begin{pmatrix}
        f(x_1) \\ \dots \\ f(x_s)
    \end{pmatrix}.
\end{equation}

Similarly, the above numerical implementation can be applied to a nonlinear integral transform
\begin{equation}\label{eqn:integraltransform2}
    (\bar{T}f)(z)=\int_{\Omega_x}\bar{K}(f(x),x,z)f(x)dx,\quad z\in \Omega_z,
\end{equation}
which is discretized as
\begin{equation}\label{eqn:numintegraltransform2}
    (\bar{T}f)(z_j)=\frac{1}{s}\sum_{i=1}^s\bar{K}(f(x_i),x_i,z_j)f(x_i),\quad z_j\in S_z.
\end{equation}
In our numerical implementation of the IAE-Net, the above nonlinear integral transform will be applied repeatedly to achieve discretization invariance.

\subsection{IAE-Net}
\label{sec:iaenet}

 The main idea of IAE-Net is to use a recursive architecture of autoencoders to perform a series of transformations: $f\rightarrow a_0\rightarrow a_1\rightarrow \dots \rightarrow a_L\rightarrow  g$, where $\{a_i\}_{i=0,\dots,L}$ are intermediate functions defined on $\Omega_a=[0,1]^{d_a}$, where $d_a$ denotes the dimension of the intermediate function $a_i$. In principle, $d_a$ can be different for different intermediate functions. But for simplicity, we keep the same $d_a$ for all intermediate functions here. Each iterative step in the IAE-Net is obtained via a block structure proposed as IAE blocks, which we denote as $\mathcal{IAE}_i$, $i=1,\dots,L$. $L$ thus defines the number of IAE blocks in IAE-Net. Finally, each IAE block consists of several IAE in parallel. In this paper, for simplicity, we consider the case where within each individual pair of functions $f$ and $g$, the two functions are discretized at the same grid points. That is, for any $f$ and $g$, $S:=S_x=S_y$, and $d:=d_x=d_y=d_a$. As we shall see, this is due to the use of skip connections (see Section \ref{sec:dense}) in the recursive architecture of IAE-Net, which enforces that the discretizations of $a_i$ %which depend on sampling grids $S_x$ and $S_y$ via two linear transforms $F(\cdot;\theta_F)$ and $G(\cdot;\theta_G)$ parametrized by $\theta_F$ and $\theta_G$, 
 need to be the same for $i=0,\dots,L$. However, IAE-Net can still be implemented for cases even when the discretization formats or domains are not the same, i.e., $S_x\neq S_y$, by removing one skip connection, e.g., the last one.

We begin by introducing the overall structure of IAE-Net. Given numerical observations $\bar{f}$ and $\bar{g}$ of $f\in\mathcal{X}$ and $g\in\mathcal{Y}$, we consider pre and post processing structures in the network to enhance the features of the input $\bar{f}$ and condense features to reconstruct $\bar{g}$. For pre processing, a pointwise linear transformation $F(\cdot;\theta_F)$ is used to raise $\bar{f}$ to $a_0$. Post processing is done using two consecutive Fourier neural operator blocks from \cite{fno} to stabilize the training followed by a pointwise linear transform to project $a_L$ to $\bar{g}$. This model is denoted as $G(\cdot;\theta_G)$. That is, $a_0(x_j)=F(\bar{f}(x_j);\theta_F)$ and $\bar{g}(y_j)=G(a_L(y_j);\theta_G)$. The overall computational flow from $\bar{f}$ to $\bar{g}$ is thus given by
\begin{equation*}
    \bar{f}\xrightarrow{F} a_0\xrightarrow{\mathcal{IAE}_1} a_1 \xrightarrow{\mathcal{IAE}_2} \dots \xrightarrow{\mathcal{IAE}_L} a_L\xrightarrow{G} \bar{g}.
\end{equation*}

The inclusion of $F(\cdot;\theta_F)$ raises the observed data to the same higher dimensional space as the intermediate layers of IAE-Net. This standardizes the width of each layer and allows the raised $a_0$ to be used in skip connections. In other words, $F$ raises the observed data to a higher dimensional space, creating a higher dimensional feature representation of the input data $\bar{f}$ that is used as skip connections to subsequent IAE blocks.

\subsubsection{Integral Autoencoders (IAE)}
\label{sec:iae}

Integral Autoencoders (IAE) form the most basic component within each IAE block. Traditional NNs assume that a fixed discretization format is used to discretize all the observed functions in $\mathcal{X}$ and $\mathcal{Y}$. In discretization invariant learning, this assumption is relaxed to allow the discretization to vary. Due to this relaxation, standard NN operations like FNNs can no longer be directly applied to functions with different discretization formats. We propose IAE as a solution to circumvent this constraint via integral transforms. 

For simplicity, we introduce an IAE that transforms an input function $a$ defined on $\Omega=[0,1]^d$ to another function $b$ defined on the same domain $\Omega$. It is easy to generalize it to the case when $b$ is defined on a different domain. Both $a$ and $b$ are discretized with $s$ grid points $S=\{x_j\}_{j=1}^s\subset\Omega$, where $s$, and hence $S$, are allowed to vary. To achieve this goal, an NN $\Phi(\cdot;\theta_{\Phi})$ with a special architecture will be constructed, where $\theta_{\Phi}$ represents the parameters of this NN. We model $\Phi(\cdot;\theta_{\Phi})$ using a three-step transformation from $a$ to $b$, following the computational flow below
\begin{equation*}
    a\rightarrow v\rightarrow u\rightarrow b=:\Phi(a;\theta_{\Phi}),
\end{equation*}
where $v$ and $u$ are intermediate functions also defined on $\Omega_z=[0,1]^d$, but discretized using grid points $S_z=\{z_j\}_{j=1}^m\subset \Omega_z$ independent of $S$, where $m$ is an integer smaller or equal to $s$. When $m$ is set to be much smaller than $s$, the transformation from $a$ to $b$ can be understood as an autoencoder for dimension reduction and function reconstruction. $S_z$ is a fixed set of grid points chosen as hyper-parameters by the user. Since $S_z$ is fixed, standard fixed-size NN like FNNs can be applied to parametrize the mapping from $v$ to $u$. This fixed-size NN is denoted as $\phi_0(\cdot;\theta_{\phi_0})$ with parameters $\theta_{\phi_0}$, and is designed using an $N$-layer FNN (see Equation \eqref{eqn:fnn}).

To map $a$ to $v$ with discretization invariance, an encoder is designed as a nonlinear integral transform with an NN $\phi_1(\cdot;\theta_{\phi_1})$, to parametrize the integral transform kernel, where $\theta_{\phi_1}$ is the set of NN parameters. Encoding is performed via a nonlinear integral transform, defined by
\begin{equation}\label{eqn:forwardintegraltransform}
    v(z)=\int_{\Omega}\phi_1(a(x),x,z;\theta_{\phi_1})a(x)dx,\quad z\in \Omega_z.
\end{equation}
There are two reasons for using integral transforms: 1) The resulting function $v$ is sampled using another set of grid points $S_z\subset \Omega_z$ chosen by the user, and thus can be chosen to be independent of any $S\subset\Omega$. This means that Equation \eqref{eqn:forwardintegraltransform} can be used to map $a$ on any $S$ in $\Omega$ to a fixed $S_z$ in $\Omega_z$. 2) The process can be efficiently implemented using matrix multiplication as discussed in Section \ref{sec:integral}, allowing for fast evaluation if the corresponding transformation matrix has hierarchical low-rank structures, which is left as interesting future work following \cite{fan2019solving,Fan2019,switchnet,Li2020}.

On the same note, to map $u$ to $b$ with discretization invariance, a decoder architecture is designed with the help of another NN $\phi_2(\cdot;\theta_{\phi_2})$ to parametrize the integral transform kernel, where $\theta_{\phi_2}$ is the set of NN parameters. Decoding is performed via another nonlinear integral transform, defined by
\begin{equation}\label{eqn:backwardintegraltransform}
    b(x)=\int_{\Omega_z}\phi_2(u(z),x,z;\theta_{\phi_2})u(z)dz,\quad x\in \Omega.
\end{equation}
This transforms $u$ from the fixed discretization with grid points $S_z\subset\Omega_z$ back to any arbitrary discretization with grid points $S\subset\Omega$. 

$a(x)$ and $u(z)$ are pointwise auxiliary information used as input to $\phi_1$ and $\phi_2$, respectively, to improve the learning performance of the two NNs. The discretization invariance of IAE-Net is still valid as the values of $a(x)$ and $u(z)$ for a single grid point $x\in S$ and $z\in S_z$ have a fixed size. Thus, the inputs to $\phi_1$ and $\phi_2$ are of fixed size and are independent of $S$ and, hence, they can be constructed using a fixed-size NN like FNNs (see Equation \eqref{eqn:fnn}). After the integral transform, additional channel MLPs (see Equation \eqref{eqn:mlp}), which does not impact the discretization invariance of the model, can be used to perform channel mixing of the outputs.
% to output the kernel values in Equations \eqref{eqn:forwardintegraltransform} and \eqref{eqn:backwardintegraltransform}.

Data-driven methods have been successfully used to learn integral kernels. \cite{CAI201489} developed a data-driven tight frame encoder and decoder to learn sparse features of images. This is applied to develop an unsupervised learning algorithm in image denoising \cite{CAI201489,Bao2015} and a supervised learning algorithm in image classification \cite{6909888,7293682}. Similarly, $\phi_1$ and $\phi_2$ could be trained to mimic classic integral transforms, e.g., the Fourier transform and the Wavelet transform. The shift from existing integral kernels to a data-driven approach is motivated by the powerful approximation capability of NNs. The use of NNs in IAE has two main advantages. First, data-driven NNs allow IAE to capture the low-dimensional structures of the problem through a supervised learning task in the same spirit of PCA with NNs as basis functions. Next, this minimizes the dependence of IAE-Net to knowledge-driven choices for the design of the integral, allowing IAE-Net to generalize to a broad range of applications as a universal solver.

%In the forward transform, $\phi_1$ is implemented with a 1 hidden layer FNN taking in the positional information $x_i$ together with auxiliary information $f(x_i)$ and outputs the corresponding kernel values $\phi_1(f(x_i),x_i,z_j;\theta_1)$. Here, $z_j$, where $j=1,\dots,m$ are points from $\Omega_z$, a fixed discretization of $m$ constant points from the intermediate finite discretization space $\mathcal{Z}$. The auxiliary information is provided as input to $\phi_1$ to provide additional information for the kernel to learn data-driven properties of the problem. The design of $\phi_2$ is similar, but instead outputs the kernel values to perform the backward integral transform from $\mathcal{Z}$ to $\mathcal{Y}$, the domain of the objective value $g(y)$.

\begin{figure}[t]
    \centering
    \includegraphics[width=0.8\columnwidth]{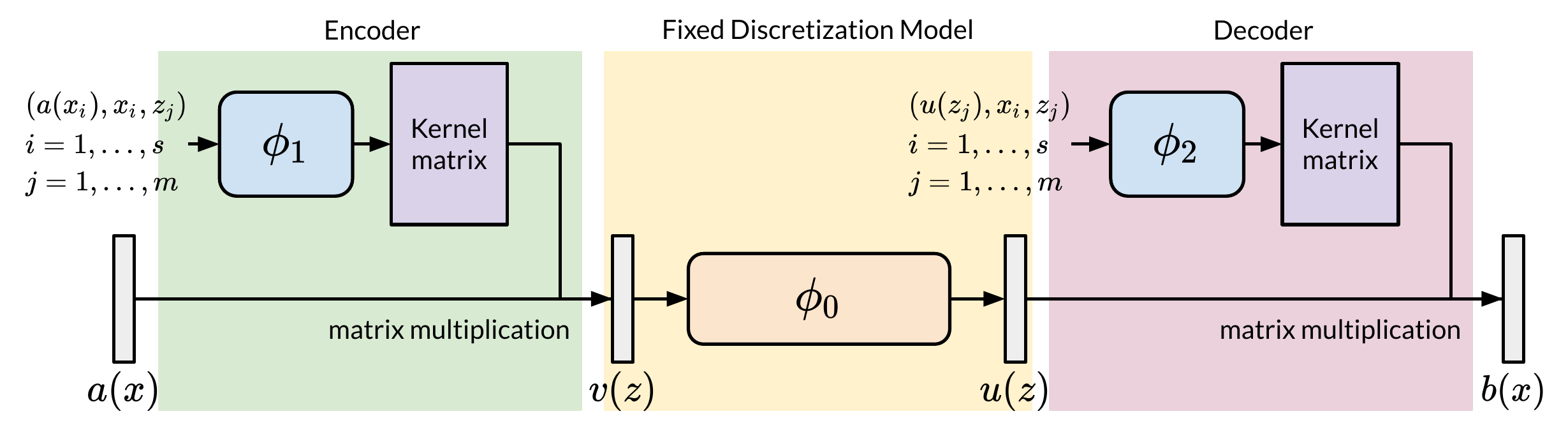}
    \caption{Models of the nonlinear integral transforms in the 3-component structure of a one-layer IAE. The encoder implements the nonlinear integral transform in Equation \eqref{eqn:forwardintegraltransform} using matrix multiplication, using $\phi_1$ to build the kernel matrix (blue dashed box). The decoder implements the nonlinear integral transform in Equation \eqref{eqn:backwardintegraltransform}, using $\phi_2$ to build the kernel matrix (blue dashed box). $\phi_0$ is an FNN described in Section \ref{sec:iae}. The kernel matrix represents the matrix $K$ (see Equation \eqref{eqn:dotproductoriginal}) with $i,j$-entries $\phi_1(a(x_i),x_i,z_j)$ and $\phi_2(u(z_i),x_j,z_i)$ respectively for the encoder and decoder.}
    \label{fig:proposedmodel}
\end{figure}

Similar to the integral kernels in classic integral transforms, the above IAE encoder and decoder in Equations \eqref{eqn:forwardintegraltransform} and \eqref{eqn:backwardintegraltransform} can be easily implemented numerically using matrix multiplication as discussed in Section \ref{sec:integral}. In summary, IAE is visualized in Figure \ref{fig:proposedmodel}.

%Using the forward and backward integral transform designed using $\phi_1$ and $\phi_2$ as the encoder and decoder ensures that IAE-Net can be evaluated at any arbitrary discretization of $f$ to obtain any target discretization of $g$. The integral transform to an independent discretization space $\Omega_z$ allows the use of FNN via $\phi_0$ on the discretization axis without violation of the discretization invariant property, otherwise impossible using the original input. The design of $\phi_1$ and $\phi_2$ as learnable NNs in contrast to a fixed choice of integral transform in existing attempts \cite{fno} allows IAE-Net to achieve better generalization performance, able to be applied to a broader range of applications, from well-posed benchmark problems to ill-posed problems in a wide range of fields.

Equations \eqref{eqn:forwardintegraltransform} and \eqref{eqn:backwardintegraltransform} can be composed repeatedly using multiple $\phi_1$ and $\phi_2$, respectively, into an $N$-layer IAE, where $N$ represents the number of compositions. Let $\phi_1^1(\cdot;\theta_{\phi_1^1}),$ $\dots,$ $\phi_1^N(\cdot;\theta_{\phi_1^N})$ denote $N$ nonlinear integral kernels. The mapping from $a$ to $v$ is implemented recursively as follows. In the $i$-th layer, for $i=1,\dots,N$, a nonlinear integral transform is performed via
\begin{equation}\label{eqn:composedforward}
    v_i(z)=\int_{\Omega_{v_{i-1}}}\phi_1^i(v_{i-1}(x),x,z;\theta_{\phi_1^i})v_{i-1}(x)dx,\quad z\in \Omega_{v_i},
\end{equation}
where $v_0:=a$, $v_i$ is defined on $\Omega_{v_i}$ for $i=0,\dots,N$. $v_N$ is denoted as $v$, thus the mapping from $a$ to $v$ follows $a\rightarrow v_1\rightarrow\dots\rightarrow v_N=v$. Similarly, let $\phi_2^1(\cdot;\theta_{\phi_2^1}),\dots,\phi_2^N(\cdot;\theta_{\phi_2^N})$ denote $N$ nonlinear integral kernels. The mapping from $u$ to $b$ is also performed recursively, such that in the $i$-th layer, where $i=1,\dots,N$, a nonlinear integral transform is performed, given by
\begin{equation}\label{eqn:composedbackward}
    u_i(x)=\int_{\Omega_{u_{i-1}}}\phi_2^i(u_{i-1}(z),x,z;\theta_{\phi_2^i})u_{i-1}(z)dz,\quad x\in \Omega_{u_i},
\end{equation}
where $u_0:=u$, $u_i$ is defined on $\Omega_{u_i}$ for $i=0,\dots,N$. $u_N$ is denoted as $b$, thus the mapping from $u$ to $b$ follows $u\rightarrow u_1\rightarrow\dots\rightarrow u_N=b$. The spaces $\Omega_{v_i}$, for $i=1,\dots,N$, and $\Omega_{u_j}$, for $j=0,\dots,N-1$, as well as their corresponding set of grid points, are chosen by the user. The set of grid points $S_z$ of $\Omega_{v_N}=\Omega_{u_0}=\Omega_z$ is chosen to be fixed and, hence, is independent of any set of grid points $S$ of $\Omega$. This allows the mapping from $v$ to $u$ to be implemented using a fixed-size FNN $\phi_0$. Aside from the integral kernels $\phi_i^j$, for $i=1,2$ and $j=1,\dots,N$, no other FNN is present in the implementation of the nonlinear integral transforms. These nonlinear integral kernels in Equations \eqref{eqn:composedforward} and \eqref{eqn:composedbackward} only require pointwise information as inputs, thus they do not depend on how the input function is discretized. Hence, except for $\Omega_{v_N}$ and $\Omega_{u_0}$, the number of grid points for the other intermediate spaces are allowed to depend on $s$, the number of grid points from $\Omega$ used to discretize $a$ and $b$.

%The final mapping, using $\phi_1^1(\cdot;\theta_{\phi_1^1})$,$\dots$,$\phi_1^N(\cdot;\theta_{\phi_1^N})$ to build the composite mapping from $a$ to $v$, $\phi_2^1(\cdot;\theta_{\phi_2^1})$,$\dots$,$\phi_2^N(\cdot;\theta_{\phi_2^N})$ 
The final mapping, using $\phi_1^i(\cdot;\theta_{\phi_1^i})$, for $i=1,\dots,N$, to build the composite mapping from $a$ to $v$, $\phi_2^j(\cdot;\theta_{\phi_2^j})$, for $j=1,\dots,N$,
to build the mapping from $u$ to $b$, and $\phi_0$ to map $v$ to $u$, forms an $N$-layer IAE. The objective of composing multiple integral transforms is to learn a feature-rich representation of the data, analogous to the multilayer MLP and FNNs. For simplicity of notation, for the rest of the paper, unless otherwise stated, we refer to the integral kernels of an IAE by considering a $1$-layer IAE, i.e., $\phi_1$ (Equation \eqref{eqn:forwardintegraltransform}) and $\phi_2$ (Equation \eqref{eqn:backwardintegraltransform}) denotes the integral kernels of an IAE.

\subsubsection{Multi-Channel Learning}\label{sec:mcl}

Instead of using the original function as the input and output of IAE (defined in Section \ref{sec:iae}), it is typically beneficial to process information in a transformed domain, where the original information admits sparsity after transformation. This is useful to capture various phenomena emphasized in other domains. For example, oscillatory phenomena, commonly found in problems like signal processing, are better represented in the Fourier domain, where highly oscillatory information is found in the higher frequency bands, and the lower frequency data are well separated from the higher frequency data. In IAE-Net, we propose a multi-channel learning framework, denoted as IAE blocks, or $\mathcal{IAE}(\cdot;\theta_{\mathcal{IAE}})$, to allow IAE-Net to solve a wider range of problems. Here $\theta_{\mathcal{IAE}}$ denotes the parameters of this block.

In this multi-channel learning framework, IAEs are applied in parallel, where in addition to the original input function, alternative branches are applied to transformed inputs using known integral transforms like the Fourier or Wavelet Transform. A $c$-channel IAE block is constructed by first denoting $\mathcal{F}_1,\dots,\mathcal{F}_{c-1}$ as $c-1$ integral transforms, together with their corresponding inverses $\mathcal{F}^{-1}_1,\dots,\mathcal{F}^{-1}_{c-1}$. These integral transforms map the input function $a$ to another transformed domain. The transformed functions $\mathcal{F}_{i}a$, $i=1,\dots,c-1$, together with $a$, become $c$ different inputs used for each of the branches (see Figure \ref{fig:parallel} for a visualization). These functions are used to implement a $c$-channel IAE block via
\begin{equation}\label{eqn:multichannel}
    \mathcal{IAE}(a;\theta_{\mathcal{IAE}}):=M([\Phi_1(a;\theta_{\Phi_1}),\mathcal{F}^{-1}_1\Phi_2(\mathcal{F}_1a;\theta_{\Phi_2}),\dots,\mathcal{F}^{-1}_{c-1}\Phi_c(\mathcal{F}_{c-1}a;\theta_{\Phi_c})];\theta_M),
\end{equation}
where $[\cdot,\dots,\cdot]$ denotes the concatenation operator along the channel axis; $\Phi_i$, for $i=1,\dots,c$, represents the $i$-th IAE; and $M(\cdot;\theta_M)$ parametrized by $\theta_M$ is a pointwise MLP (see Equation \eqref{eqn:mlp}) post-processing and merging the outputs of each channel. Figure \ref{fig:parallel} shows a $2$-channel IAE block designed using the original input and the Fourier transform. Appendix \ref{apd:comparison_parallel_blocks} and \ref{apd:transformer_architecture} provides additional numerical experiments motivating the choice of using a multi-channel learning framework and a channel MLP for post processing.

%Given the outputs of each parallel IAE branch, an additional width axis operation is used to perform merging and post processing. This width axis operation is designed using a width/channel axis multilayer perceptron (MLP) consisting of 1 hidden layer. The choice of a channel MLP ensures that the operations are performed on the constant width axis, allowing IAE-Net to retain it's discretization invariant structure, while performing further post processing of the outputs. A multi-channel IAE, together with this channel MLP, constutes our final IAE block, and this structure is shown in Figure \ref{fig:blockstructure}. Appendix \ref{apd:transformer_architecture} shows that the use of a channel MLP to perform post-processing of the parallel IAE branches allows IAE-Net to perform almost 7 times better on the benchmark data set.

\subsubsection{Densely Connected Multi-Block Structure}
\label{sec:dense}

The IAE blocks defined in Section \ref{sec:mcl} are composed repeatedly with pointwise linear transforms of the outputs from previous IAE blocks to form a densely connected multi-block structure motivated by DenseNet \cite{densenet}. This follows the computational flow introduced in Section \ref{sec:iaenet} and will be proposed as the final IAE-Net, denoted as $\Psi^n(\cdot;\theta_{\Psi^n})$. Here, $\theta_{\Psi^n}$ denotes the parameters of $\Psi^n$.

\begin{figure}[t]
    \centering
    \includegraphics[width=0.8\columnwidth]{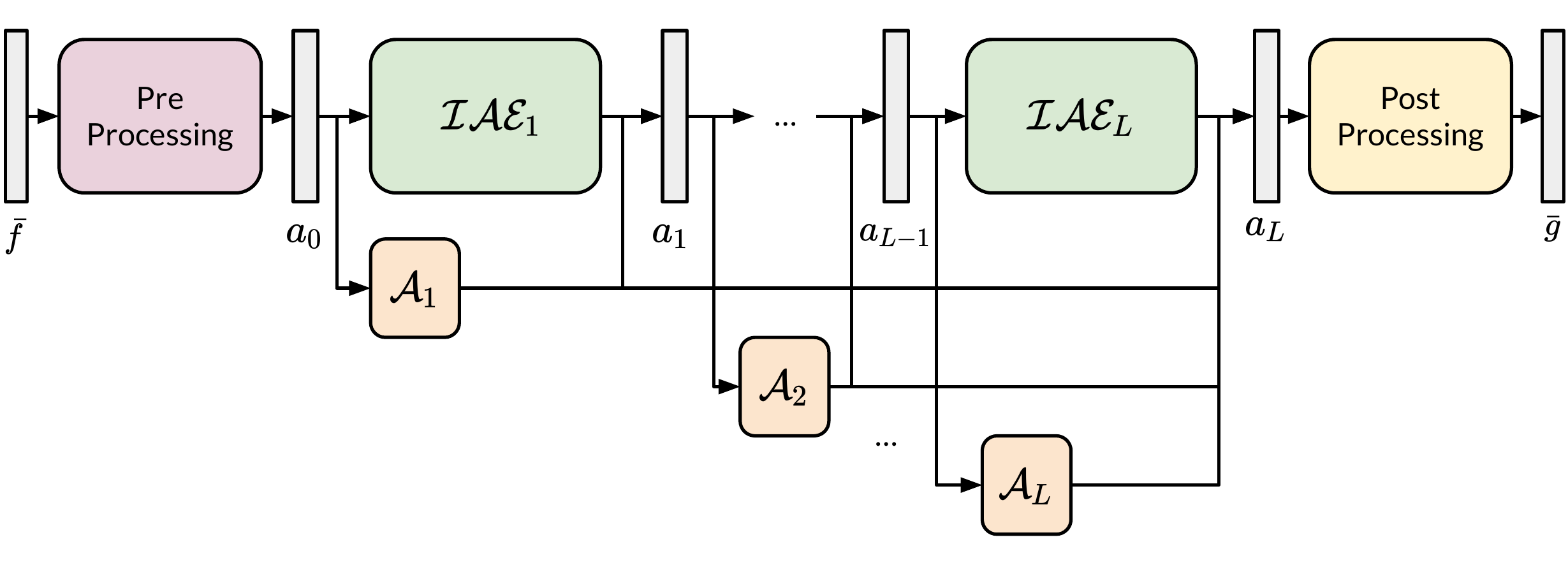}
    \caption{Structure of a $L$ block IAE-Net. For $i=1,\dots,L$, $\mathcal{IAE}_i$ represents IAE blocks, described in Equation \eqref{eqn:multichannel}, $\mathcal{A}_i$ refers to the pointwise linear transforms implemented in a densely connected structure in Equation \eqref{eqn:densenet}, and the pre and post processing block is built with $F$ and $G$ described in Section \ref{sec:iaenet}. The numerical implementation of all the components are described in Section \ref{sec:training_details}.}
    \label{fig:comparison_structure}
\end{figure}

The composition of the blocks in IAE-Net is formulated as follows. Consider the $i$-th IAE block in a $L$-block IAE-Net, denoted by $\mathcal{IAE}_i(\cdot;\theta_{\mathcal{IAE}_i})$, where $i=1,\dots,L$. In addition, $\mathcal{A}_j(\cdot;\theta_{\mathcal{A}_j})$ is defined as pointwise affine transforms parametrized by $\theta_{\mathcal{A}_j}$, taking in $a_j$ as its input, for $j=0,\dots,L-1$. Then, this composition is performed with skip connections from all the preceding outputs $a_0,\dots,a_{i-1}$ via
\begin{equation}\label{eqn:densenet}
    a_i=\sigma(\mathcal{IAE}_i(\mathcal{M}_i([\mathcal{A}_0(a_0;\theta_{\mathcal{A}_0}),\mathcal{A}_1(a_1;\theta_{\mathcal{A}_1}),\dots,\mathcal{A}_{i-1}(a_{i-1};\theta_{\mathcal{A}_{i-1}})];\theta_{\mathcal{M}_i});\theta_{\mathcal{IAE}_i})),
\end{equation}
where $[\cdot,\dots,\cdot]$ denotes the concatenation operator along the channel axis; $\mathcal{M}_j(\cdot;\theta_{\mathcal{M}_j})$ parametrized by $\theta_{\mathcal{M}_j}$ is a pointwise MLP (see Equation \eqref{eqn:mlp}) merging the outputs of each of the linear transformed outputs from the previous IAE blocks; and $\sigma$ is an activation function. The resulting IAE-Net is visualized in Figure \ref{fig:comparison_structure}.

This densely connected structure diversifies the features of the inputs to each IAE block, allowing IAE-Net to learn complex representations of the data. In addition, it helps to lessen the gradient vanishing issue commonly observed in FNNs \cite{resnet}, thus providing additional stability to IAE-Net. This is shown through extensive experiments in Section \ref{sec:experiments}, where IAE-Net achieves consistent performance better than numerous other deep NN structures in a wide range of applications. Appendix \ref{apd:number_of_blocks} presents numerical experiments to show that the performance of IAE-Net is almost improved by $100\%$ as compared to using a single block.

\subsubsection{Data Driven Discretization Invariant Learning}
\label{sec:data_augment}

Let $\Psi^n(\cdot;\theta_{\Psi^n})$ denote the proposed IAE-Net parametrized by $\theta_{\Psi^n}$ (defined in Section \ref{sec:dense}). For simplicity, we assume that $\Psi^n$ is built using $1$-layer IAEs within each IAE block, so $\phi_1(\cdot;\theta_{\phi_1})$ (see Equation \eqref{eqn:forwardintegraltransform}) and $\phi_2(\cdot;\theta_{\phi_2})$ (see Equation \eqref{eqn:backwardintegraltransform}) refers to the integral kernels used in a $1$-layer IAE located in $\Psi^n$, where $\theta_{\phi_1}$ and $\theta_{\phi_2}$ parametrize $\phi_1$ and $\phi_2$ respectively. Given that these integral kernels are modelled as NNs, their performance depends on the given data. Take, for example, a data set consisting of data sampled from a specific set of grid points $S$ in $\Omega$. During training, $\phi_1$ and $\phi_2$ are only trained to output suitable kernel values for the grid points in $S$. The performance of these kernels deteriorates when tested on other grid points not found in $S$. Thus, being data-driven, $\phi_1$ and $\phi_2$ suffer from poor generalization errors. To reduce this error, this paper proposes a novel data augmentation procedure to exploit the discretization invariant structure of IAE-Net. The procedure involves augmenting the training process with different resolutions of the data, obtained via interpolation of the given data. This method promotes the learning of $\phi_1$ and $\phi_2$, allowing them to output suitable kernel values from inputs of various sizes.

Let $\mathcal{X}$ denote a function space of functions defined on $\Omega$, and define $X_{S}=\{\bar{f}|f\in\mathcal{X}\}$ as the set of numerical observations $\bar{f}$ of functions $f\in\mathcal{X}$ sampled using grid points in $S\subset\Omega$. Typically, during training, the available training data are sampled from $X_S$. Let $\mathcal{S}$ denote the set of all possible non-empty finite subsets of $\Omega$. i.e., $S\subset\mathcal{S}$ represents a set of grid points from $\Omega$. Given observations of a problem sampled from grid points $S\in\mathcal{S}$, and $T\in\mathcal{S}$ another set of grid points of $\Omega$, an interpolator function $I_{T}:X_{S}\rightarrow X_{T}$ maps numerical observations $\bar{f}\in X_{S}$ to numerical observations $I_{T}(\bar{f})\in X_{T}$ sampled (using grid points in $T$) from the interpolant $l\in\mathcal{X}$ obtained from $\bar{f}$. By definition, an interpolant $l\in\mathcal{X}$ of $f$ from observations $\bar{f}\in X_S$ is a function approximating $f$ that agrees with $f$ on the set of points in $S$, i.e., $\bar{l}=\bar{f}$. Typical examples of interpolants are linear, bilinear, and cubic spline interpolation.

Define $\mathcal{I}:=\{I_{T}|T\in\mathcal{S}\}$ as the set of all such interpolator functions. At each iteration of the SGD when training IAE-Net, the training samples are augmented with the interpolations of the data using $I_T\in \mathcal{I}$, i.e., we optimize the following objective function via SGD:
\begin{equation}\label{eqn:loss}
    \min_{\theta_{\Psi^n}}\mathbb{E}_{(\bar{f},\bar{g})\sim p_{data}}\mathbb{E}_{I_T\sim \mathcal{I}}[L(\Psi^n(\bar{f};\theta_{\Psi^n}), \bar{g})+\lambda L(\Psi^n(I_T(\bar{f});\theta_{\Psi^n}), I_T(\bar{g}))],
\end{equation}
where $\lambda$ is a hyperparameter, $p_{data}$ is the distribution of the training data pair $(\bar{f},\bar{g})$, and $\Psi^n(\cdot;\theta_{\Psi^n})$ denotes the proposed IAE-Net parameterized by $\theta_{\Psi^n}$. Equation \eqref{eqn:loss} leverages the discretization invariance of IAE-Net to train the integral kernels of each IAE within $\Psi^n$ with inputs of varying discretizations. This is crucial to allow IAE-Net to achieve discretization invariance even with data-driven kernels. Usually, $\mathcal{S}$ contains an infinite number of discretization formats. So in practice, only a finite subset $U\subset\mathcal{S}$ of commonly used discretization formats is considered.

\section{Numerical Experiments}
\label{sec:experiments}

In this section, we present the experimental results comparing the performance of IAE-Net in a wide range of applications.

\subsection{Baseline Models}

The performance evaluation of IAE-Net is compared against the following baseline models.

\begin{enumerate}
    \item \textbf{ResNet+Interpolation} implements a naive method to achieve discretization invariance for fixed-discretization models. Interpolation is a simple and commonly used method for heterogeneous data structures traditionally in the literature. This method is implemented with an interpolation function interpolating the data from any arbitrary discretization to a fixed discretization $\Omega$ for input to the NN. The output performs interpolation back to $\Omega$. For this baseline, a ResNet \cite{resnet} is adopted with similar number of parameters to IAE-Net as the fixed-discretization model.
    \item \textbf{Unet} \cite{unet} is a popular tool for signal separation tasks. Although the fully convolutional structure allows Unet to take inputs sampled using different discretization formats, the model is usually used with a fixed discretization format of the input space. This is because the convolutional layers in Unet perform local processing of the inputs using a small kernel, which makes it difficult to adapt to varying input scales. This can be seen in the numerical experiments, where the performance of Unet collapses significantly when applied naively to other resolutions.
    \item \textbf{DeepONet} \cite{chenchen,deeponet} is proposed as an operator learner which is semi-discretization invariant. The original model, although discretization invariant to the output, is however not discretization invariant to the inputs, where a fixed discretization is used to learn an encoding of the input function.
    \item \textbf{Fourier Neural Operator (FNO)} \cite{fno} is proposed as a seminal discretization invariant operator learner on numerous mathematically well-posed benchmark data sets.
    \item \textbf{Fourier Transformer (FT), Galerkin Transformer (GT)} \cite{galerkintransformer} are proposed as discretization invariant operators applying transformer encoders to encode given data \cite{attention} and the blocks in \cite{fno} for post-processing. The model achieves state-of-the-art performance in its benchmark data sets.
\end{enumerate}

The original code provided by the authors is referenced for all the baselines, obtained from their Github pages. For the experiments, their choice of settings for the hyperparameters and training procedures are used (e.g., modes, width in FNO, and initialization for attention matrices in FT, GT) to train their models. Since DeepONet is not discretization invariant to the input data, to obtain results for different resolutions, the models are separately trained using different resolutions. Similarly, Unet requires repeated training when test data have different resolutions. In the next section, the numerical setup of the experiments will be provided in detail. Following this, the remainder of Section \ref{sec:experiments} will show the numerical experiments of IAE-Net and the above baselines in a wide range of problems. We will numerically demonstrate the following conclusions: 
\begin{itemize}
    \item Naive interpolation-based method with a fixed-size neural network, FNO, GT/FT, and IAE-Net can achieve zero/few shot generalization in well-posed problems like solving parametric PDEs and initial value problems.
    \item Existing methods fail to provide zero/few shot generalization in ill-posed problems and/or for highly oscillatory data, while IAE-Net succeeds to generalize well.
    \item IAE-Net outperforms all existing baseline methods with better accuracy in all test examples. %In ill-posed problems, IAE-Net offers powerful zero/few shot generalization capability and can perform well even when all existing methods fail. In well-posed problems, existing discretization invariant methods, the naive ResNet with interpolation, and IAE-Net all admit zero/few shot generalization capacity, but IAE-Net provides the best accuracy in all test examples.
\end{itemize}

\subsection{Training Details}
\label{sec:training_details}

For all applications, grid points are sampled using a uniform grid in $[0,1]^d$, where $d$ represents the dimension of the problem. Let $S_1\subset [0,1]$ and $S_2\subset [0,1]^2$, denote the set of grid sizes used to discretize the input and output functions for the $1$-d and $2$-d problem respectively. For $1$-d problems, i.e., $d=d_x=d_y=1$, a uniform grid sampled from $[0,1]$ is used. That is, $S_1=\{\frac{i-1}{s-1}\}^s_{i=1}$ where $s$ denotes the discretization size, which is not fixed. For the fixed $S_z$ containing the grid points for discretizing the intermediate function $u$ and $v$ in IAE (see Section \ref{sec:iae}), we set $m=256$, where $m$ denotes the discretization size, thus $S_z=\{\frac{i-1}{255}\}^m_{i=1}$. For the $2$-d problem, i.e., $d=d_x=d_y=2$, a $2$-d uniform mesh is used, such that $S_2=S_1\times S_1=\{(\frac{i-1}{s-1},\frac{j-1}{s-1})\}$, with $\times$ representing the cartesian product, and $i,j=1,\dots,s$ where $s$ denotes the discretization size along one axis. For $S_z$ in the $2$-d problem, we set $m=64$, where $m$ denotes the discretization size along one axis, thus $S_z=\{(\frac{i-1}{63},\frac{j-1}{63})\}$, with $i,j=1,\dots,64$.

The recursive architecture in IAE-Net is constructed by stacking four $\mathcal{IAE}$'s in the densely connected structure described in Equation \eqref{eqn:densenet}, with the ReLU activation function as $\sigma$, and a single hidden layer pointwise MLP for $\mathcal{M}$. Thus $L=4$ in the experiments. The pointwise affine transform $\mathcal{A}$ in Equation \eqref{eqn:densenet} is implemented using pointwise linear transforms. Each $\mathcal{IAE}$ is constructed following Equation \eqref{eqn:multichannel}, using a $2$-channel IAE block taking in the original input and the Fourier transformed input for each of the respective channels, visualized in Figure \ref{fig:parallel}.

For the $1$-d case, a $1$-layer IAE is used, while a $2$-layer IAE is used in the $2$-d case. For the nonlinear integral transforms $\phi_1^1$ and $\phi_1^2$ (Equation \eqref{eqn:composedforward}) in the $2$-layer IAE, $\phi_1^1$ is used to first map $S_2$ to $S_f=\{(\frac{i-1}{63},\frac{j-1}{s-1})\}$, where $i=1,\dots,m$ and $j=1,\dots,s$ before mapping to $S_z$ using $\phi_1^2$, following Equation \eqref{eqn:composedforward}. This can be understood as performing an integral transform on the $x$-axis, followed by the $y$-axis. For the nonlinear transforms $\phi_2^1$ and $\phi_2^2$, $\phi_2^1$ is used to first map $S_z$ to $S_f$, before mapping to $S_2$ using $\phi_2^2$, following Equation \eqref{eqn:composedbackward}. A single hidden layer FNN is used to design $\phi_1$ and $\phi_2$ in the $1$-d case, as well as $\phi_1^1$, $\phi_1^2$, $\phi_2^1$ and $\phi_2^2$ in the $2$-d case. A single hidden layer pointwise MLP is applied to the outputs of the nonlinear integral transforms. A two hidden layer FNN is used for $\phi_0$ in both cases. $M$ (see Equation \eqref{eqn:multichannel}) is implemented using a single hidden layer pointwise MLP. As defined in Section \ref{sec:iaenet}, pre processing is done using a pointwise linear transform $F$ mapping the input observations $\bar{f}$ to $\mathbb{R}^w$, where $w$ denotes the length of the projected vector. Post processing is done using two FNO blocks followed by a pointwise linear transform, following $G$ in Section \ref{sec:iaenet} mapping $a_L$ to $\bar{g}$. We set $w=64$ for both the $1$-d and $2$-d problems. Figure \ref{fig:comparison_structure} shows the structure of IAE-Net.

The data augmentation (DA) strategy proposed in Equation \eqref{eqn:loss} is used by default for IAE-Net, with $\lambda=1$, using interpolation of the training data to the testing sizes. Lower resolution data representing $I_T(\bar{f})$ and $I_T(\bar{g})$ are obtained by downsampling from a higher resolution, while higher resolution data are obtained with cubic and bicubic interpolation for the $1$-d and $2$-d problems respectively. IAE-Net is trained for all the applications with a learning rate of $0.001$, using Adam optimizer as the choice of optimizer. A batch size of 50 for $1$-d problems, and 5 for $2$-d problems, is used, and training is performed for 500 epochs using a learning rate scheduler scaling the learning rate by $0.5$ on a plateau with patience of $20$ epochs. All the models are trained using the $L2$ relative error, denoted as $\mathcal{L(\cdot,\cdot)}$, as the loss function, defined by 
\begin{equation}\label{eqn:l2loss}
    \mathcal{L}(x,y):=\frac{||x-y||^2}{||y||^2}.
\end{equation}
The performance of all the models will also be evaluated using the mean $L2$ relative error, defined in the above Equation \eqref{eqn:l2loss}. The experiments are run with a 48GB Quadro RTX 8000 GPU.

Alternative architectures for IAE-Net are considered for additional baselines. The first adopts the ResNet \cite{resnet} style skip connection performing an elementwise summation of the skip connection with the output of the IAE-Net block. For this baseline, instead of Equation \eqref{eqn:densenet}, the output for the $i$-th IAE block is defined as
\begin{equation}\label{eqn:resnet}
    f_i=\sigma(\mathcal{IAE}_i(f_{i-1})+\mathcal{A}_i(f_{i-1})),
\end{equation}
with $\sigma$ as the ReLU activation, and $\mathcal{A}_i$ as a pointwise linear transform. The second is a base model without any skip connections between the blocks. These baselines evaluate the performance of IAE-Net with varied forms of skip connections and motivates our choice of using the densely connected structure in the proposed IAE-Net. The two models are denoted IAE-Net (ResNet) and IAE-Net (No Skip) respectively.

\subsection{Predictive Modelling}
\label{sec:burgers}

The first application lies in the solving of predictive modelling problems. For this problem, the Burgers Equation is considered, given by
\begin{equation}\label{eqn:burgers}
    \begin{split}
        \partial_tu(x,t)+\partial_x(u^2(x,t)/2)&=\nu\partial_{xx}u(x,t),\quad x\in(0,1),t\in(0,1]\\
        u(x,0)&=u_0(x).
    \end{split}
\end{equation}
with periodic boundary conditions. In this application, the objective is to learn the mapping from an initial state/condition $u_0$ to the solution $u$ at time $t=1$. That is, the operator to be learnt is $\Psi$ mapping $u_0(.,0)\rightarrow u(.,1)$ for $x\in(0,1)$. Here, $\nu$ represents the viscosity of the problem. This problem is used as one of the benchmark problems in \cite{fno, galerkintransformer}.

The initial conditions to generate this data set, provided by \cite{fno} in \url{https://github.com/zongyi-li/fourier_neural_operator}, is generated according to $u_0\sim\mu$, where $\mu=\mathcal{N}(0,\sigma^2(-\Delta+\tau^2I)^{-\alpha})$, using $\tau=5$, $\sigma=\tau^2$ and $\alpha=2$. Viscosity is set to $\nu=0.1$, and the equation is solved via a split step method where the heat equation component is solved exactly in the Fourier space and the non-linear part is advanced in the Fourier space, using a fine forward Euler method. A total of 10000 training samples and 1000 testing samples are generated.

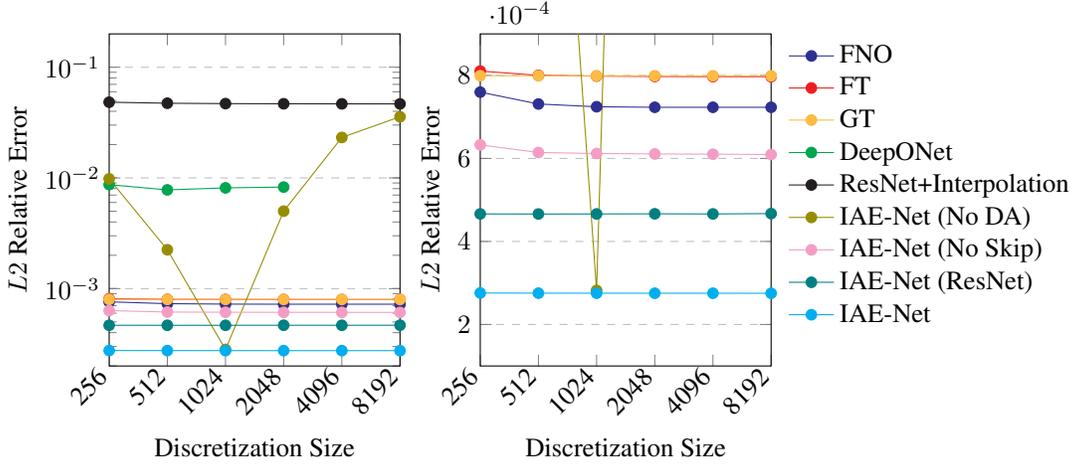
\begin{figure}[t]
    \centering
    \begin{tikzpicture}
    \begin{semilogyaxis}[
        width=0.33\columnwidth,
        xlabel={Discretization Size},
        ylabel={$L2$ Relative Error},
        xmin=1, xmax=6,
        ymin=0.0002, ymax=0.2,
        xtick={1,2,3,4,5,6},
        xticklabels={256,512,1024,2048,4096,8192},
        x tick label style={
            rotate=45,
            anchor=east,
        },
        ymajorgrids=true,
        grid style=dashed,
    ]
    
\addplot[color=Blue,mark=*,]
    coordinates {(1,0.0007597)(2,0.0007311)(3,0.0007245)(4,0.0007232)(5,0.0007232)(6,0.0007232) 
};\addlegendentry{FNO}; 

\addplot[color=Red,mark=*,]
    coordinates {(1,0.0008104)(2,0.0008002)(3,0.0007983)(4,0.0007974)(5,0.0007971)(6,0.0007969) 
};\addlegendentry{FT}; 

\addplot[color=Dandelion,mark=*,]
    coordinates {(1,0.0007992)(2,0.0007991)(3,0.000799)(4,0.000799)(5,0.000799)(6,0.000799) 
};\addlegendentry{GT}; 

\addplot[color=Green,mark=*,]
    coordinates {(1,0.008695)(2,0.007797)(3,0.00813)(4,0.008265) 
};\addlegendentry{DeepONet}; 

\addplot[color=Black,mark=*,]
    coordinates {(1,0.04831)(2,0.04728)(3,0.0469)(4,0.04674)(5,0.04667)(6,0.04663) 
};\addlegendentry{ResNet+Interpolation}; 

\addplot[color=olive,mark=*,]
    coordinates {(1,0.009813)(2,0.002244)(3,0.0002821)(4,0.005006)(5,0.0232)(6,0.03567) 
};\addlegendentry{IAE-Net (No DA)}; 

\addplot[color=Lavender,mark=*,]
    coordinates {(1,0.0006326)(2,0.0006143)(3,0.0006119)(4,0.0006108)(5,0.0006102)(6,0.0006094) 
};\addlegendentry{IAE-Net (No Skip)}; 

\addplot[color=teal,mark=*,]
    coordinates {(1,0.0004664)(2,0.0004661)(3,0.0004663)(4,0.0004666)(5,0.0004663)(6,0.0004671) 
};\addlegendentry{IAE-Net (ResNet)}; 

\addplot[color=Cyan,mark=*,]
    coordinates {(1,0.0002758)(2,0.0002754)(3,0.0002753)(4,0.0002752)(5,0.0002752)(6,0.0002751) 
};\addlegendentry{IAE-Net};
    
    \legend{}
    \end{semilogyaxis}
    \end{tikzpicture}%
    \begin{tikzpicture}
    \begin{axis}[
        width=0.33\columnwidth,
        xlabel={Discretization Size},
        ylabel={$L2$ Relative Error},
        xmin=1, xmax=6,
        ymin=0.0001, ymax=0.0009,
        xtick={1,2,3,4,5,6},
        xticklabels={256,512,1024,2048,4096,8192},
        x tick label style={
            rotate=45,
            anchor=east,
        },
        ytick={0.0002,0.0004,0.0006,0.0008},
        ymajorgrids=true,
        grid style=dashed,
        legend cell align={left},
        legend style={draw=none},
        legend pos= outer north east,
    ]
    
\addplot[color=Blue,mark=*,]
    coordinates {(1,0.0007597)(2,0.0007311)(3,0.0007245)(4,0.0007232)(5,0.0007232)(6,0.0007232) 
};\addlegendentry{FNO}; 

\addplot[color=Red,mark=*,]
    coordinates {(1,0.0008104)(2,0.0008002)(3,0.0007983)(4,0.0007974)(5,0.0007971)(6,0.0007969) 
};\addlegendentry{FT}; 

\addplot[color=Dandelion,mark=*,]
    coordinates {(1,0.0007992)(2,0.0007991)(3,0.000799)(4,0.000799)(5,0.000799)(6,0.000799) 
};\addlegendentry{GT}; 

\addplot[color=Green,mark=*,]
    coordinates {(1,0.008695)(2,0.007797)(3,0.00813)(4,0.008265) 
};\addlegendentry{DeepONet}; 

\addplot[color=Black,mark=*,]
    coordinates {(1,0.04831)(2,0.04728)(3,0.0469)(4,0.04674)(5,0.04667)(6,0.04663) 
};\addlegendentry{ResNet+Interpolation}; 

\addplot[color=olive,mark=*,]
    coordinates {(1,0.009813)(2,0.002244)(3,0.0002821)(4,0.005006)(5,0.0232)(6,0.03567) 
};\addlegendentry{IAE-Net (No DA)}; 

\addplot[color=Lavender,mark=*,]
    coordinates {(1,0.0006326)(2,0.0006143)(3,0.0006119)(4,0.0006108)(5,0.0006102)(6,0.0006094) 
};\addlegendentry{IAE-Net (No Skip)}; 

\addplot[color=teal,mark=*,]
    coordinates {(1,0.0004664)(2,0.0004661)(3,0.0004663)(4,0.0004666)(5,0.0004663)(6,0.0004671) 
};\addlegendentry{IAE-Net (ResNet)}; 

\addplot[color=Cyan,mark=*,]
    coordinates {(1,0.0002758)(2,0.0002754)(3,0.0002753)(4,0.0002752)(5,0.0002752)(6,0.0002751) 
};\addlegendentry{IAE-Net};
    
    \end{axis}
    \end{tikzpicture}%
    \caption{$L2$ relative error (Equation \eqref{eqn:l2loss}) on the burgers data set with $\nu=1e^{-1}$ (Left) and its closeup (Right). Models are trained with $s=1024$ and tested on the other resolutions.}
    \label{fig:burgersbenchmark}
\end{figure}

For the experiments, the length $s=1024$ data is used to train IAE-Net, FNO, FT, GT and ResNet+Interpolation, and the resulting trained model is tested on resolutions $s=256,512,1024,2048,8192$. For DeepONet, due to the input being non-discretization invariant, the model is trained separately on the different resolutions.  Due to out-of-memory issues when loading the DeepONet data set for larger resolutions, the results for those resolutions are omitted. This particular issue highlights the drawback of non-discretization invariant models: These models require expensive re-training to train a new NN for different discretization formats, often which, the computational costs to train the model for large resolution data is extremely high. The results for this data set are shown in Figure \ref{fig:burgersbenchmark}.

In this data set, IAE-Net performs almost 3 times better than FNO, FT, and GT, while significantly outperforming both DeepONet and ResNet+Interpolation. Comparing the results of IAE-Net (No Skip) and IAE-Net (ResNet) with IAE-Net, the use of a densely connected structure significantly improves the capability of the model, performing almost 2 times better than the ResNet counterpart. This comes at minimal cost to the evaluation speed of the network, which is compared in Appendix \ref{apd:comparison_performance}. The effect of the proposed data augmentation method is seen by comparing the results of IAE-Net (No DA) to IAE-Net. In the former, IAE-Net (No DA) achieves good performance only on the resolution it is trained on, but the model does not perform well on other resolutions. In the latter, the use of data augmentation leads to consistent test error across the different resolutions. This supports the use of the proposed data augmentation, as the data reliant IAE kernels require training on different resolution scales to learn heterogeneous structures across resolutions.

\begin{figure}[t]
    \centering
    \begin{tikzpicture}
    \begin{semilogyaxis}[
        width=0.33\columnwidth,
        xlabel={Discretization Size},
        ylabel={$L2$ Relative Error},
        xmin=1, xmax=6,
        ymin=0.06, ymax=0.5,
        log basis y=10,
        xtick={1,2,3,4,5,6},
        xticklabels={256,512,1024,2048,4096,8192},
        x tick label style={
            rotate=45,
            anchor=east,
        },
        ymajorgrids=true,
        grid style=dashed,
    ]
    
\addplot[color=Blue,mark=*,]
    coordinates {(1,0.06885)(2,0.07112)(3,0.0729)(4,0.07259)(5,0.07258)(6,0.07257) 
};\addlegendentry{FNO}; 

\addplot[color=Red,mark=*,]
    coordinates {(1,0.08801)(2,0.09543)(3,0.09574)(4,0.09562)(5,0.09549)(6,0.09555) 
};\addlegendentry{FT}; 

\addplot[color=Dandelion,mark=*,]
    coordinates {(1,0.09791)(2,0.1058)(3,0.1066)(4,0.1065)(5,0.1065)(6,0.1065) 
};\addlegendentry{GT}; 

\addplot[color=Green,mark=*,]
    coordinates {(1,0.4007)(2,0.3907)(3,0.3889)(4,0.3813) 
};\addlegendentry{DeepONet}; 

\addplot[color=Black,mark=*,]
    coordinates {(1,0.3115)(2,0.3097)(3,0.3086)(4,0.3081)(5,0.3079)(6,0.3078) 
};\addlegendentry{ResNet+Interpolation}; 

\addplot[color=Lavender,mark=*,]
    coordinates {(1,0.08787)(2,0.0937)(3,0.09155)(4,0.09121)(5,0.09079)(6,0.09043) 
};\addlegendentry{IAE-Net (No Skip)}; 

\addplot[color=teal,mark=*,]
    coordinates {(1,0.07325)(2,0.07682)(3,0.07774)(4,0.07791)(5,0.07772)(6,0.07737) 
};\addlegendentry{IAE-Net (ResNet)}; 

\addplot[color=Cyan,mark=*,]
    coordinates {(1,0.06599)(2,0.06826)(3,0.06909)(4,0.0691)(5,0.06914)(6,0.06918) 
};\addlegendentry{IAE-Net};
    
    \legend{}
    \end{semilogyaxis}
    \end{tikzpicture}%
    \begin{tikzpicture}
    \begin{axis}[
        width=0.33\columnwidth,
        xlabel={Discretization Size},
        ylabel={$L2$ Relative Error},
        xmin=1, xmax=6,
        ymin=0.06, ymax=0.11,
        xtick={1,2,3,4,5,6},
        xticklabels={256,512,1024,2048,4096,8192},
        x tick label style={
            rotate=45,
            anchor=east,
        },
        ytick={0.07,0.08,0.09,0.10},
        scaled y ticks=base 10:{2},
        ymajorgrids=true,
        grid style=dashed,
        legend cell align={left},
        legend style={draw=none},
        legend pos= outer north east,
    ]
    
\addplot[color=Blue,mark=*,]
    coordinates {(1,0.06885)(2,0.07112)(3,0.0729)(4,0.07259)(5,0.07258)(6,0.07257) 
};\addlegendentry{FNO}; 

\addplot[color=Red,mark=*,]
    coordinates {(1,0.08801)(2,0.09543)(3,0.09574)(4,0.09562)(5,0.09549)(6,0.09555) 
};\addlegendentry{FT}; 

\addplot[color=Dandelion,mark=*,]
    coordinates {(1,0.09791)(2,0.1058)(3,0.1066)(4,0.1065)(5,0.1065)(6,0.1065) 
};\addlegendentry{GT}; 

\addplot[color=Green,mark=*,]
    coordinates {(1,0.4007)(2,0.3907)(3,0.3889)(4,0.3813) 
};\addlegendentry{DeepONet}; 

\addplot[color=Black,mark=*,]
    coordinates {(1,0.3115)(2,0.3097)(3,0.3086)(4,0.3081)(5,0.3079)(6,0.3078) 
};\addlegendentry{ResNet+Interpolation}; 

\addplot[color=Lavender,mark=*,]
    coordinates {(1,0.08787)(2,0.0937)(3,0.09155)(4,0.09121)(5,0.09079)(6,0.09043) 
};\addlegendentry{IAE-Net (No Skip)}; 

\addplot[color=teal,mark=*,]
    coordinates {(1,0.07325)(2,0.07682)(3,0.07774)(4,0.07791)(5,0.07772)(6,0.07737) 
};\addlegendentry{IAE-Net (ResNet)}; 

\addplot[color=Cyan,mark=*,]
    coordinates {(1,0.06599)(2,0.06826)(3,0.06909)(4,0.0691)(5,0.06914)(6,0.06918) 
};\addlegendentry{IAE-Net};
    
    \end{axis}
    \end{tikzpicture}%
    \caption{$L2$ relative error (Equation \eqref{eqn:l2loss}) on the burgers data set with $\nu=1e^{-4}$ (Left) and its closeup (Right). The closeup of the figure is shown on the right. Models are trained with $s=1024$ and tested on the other resolutions.}
    \label{fig:burgersinviscid}
\end{figure}
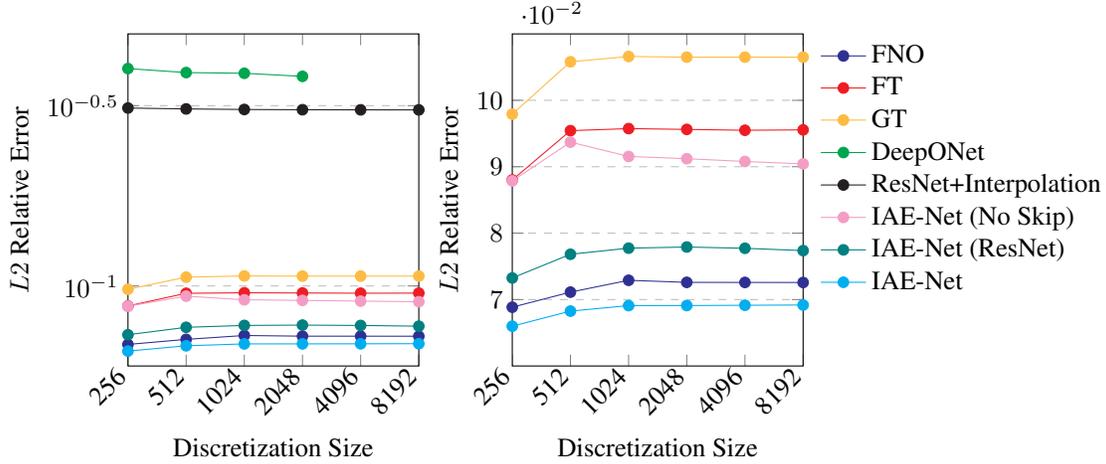

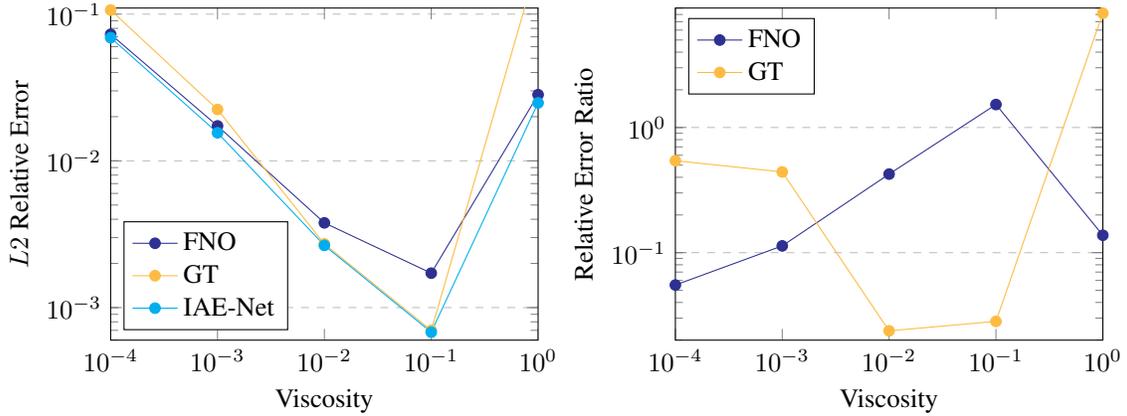
\begin{figure}[t]
    \centering
    \begin{tikzpicture}
    \begin{loglogaxis}[
        width=0.44\columnwidth,
        xlabel={Viscosity},
        ylabel={$L2$ Relative Error},
        xmin=0.0001, xmax=1,
        ymin=0.0006, ymax=0.11,
        xtick={0.0001,0.001,0.01,0.1,1},
        log basis x=10,
        % ytick={0.07,0.08,0.09,0.10},
        % scaled y ticks=base 10:{2},
        ymajorgrids=true,
        grid style=dashed,
        legend cell align={left},
        legend pos= south west,
    ]
    
\addplot[color=Blue,mark=*,]
    coordinates {(0.0001,0.0729)(0.001,0.0173)(0.01,0.00378)(0.1,0.001717)(1,0.02825) 
};\addlegendentry{FNO}; 

\addplot[color=Dandelion,mark=*,]
    coordinates {(0.0001,0.1066)(0.001,0.0224)(0.01,0.002717)(0.1,0.000699)(1,0.228) 
};\addlegendentry{GT}; 

\addplot[color=Cyan,mark=*,]
    coordinates {(0.0001,0.06909)(0.001,0.01554)(0.01,0.002654)(0.1,0.0006798)(1,0.02483) 
};\addlegendentry{IAE-Net};
    
    \end{loglogaxis}
    \end{tikzpicture}%
    \begin{tikzpicture}
    \begin{loglogaxis}[
        width=0.44\columnwidth,
        xlabel={Viscosity},
        ylabel={Relative Error Ratio},
        xmin=0.0001, xmax=1,
        ymin=0.02, ymax=9,
        xtick={0.0001,0.001,0.01,0.1,1},
        log basis x=10,
        % ytick={0.07,0.08,0.09,0.10},
        % scaled y ticks=base 10:{2},
        ymajorgrids=true,
        grid style=dashed,
        legend cell align={left},
        legend pos= north west,
    ]
    
\addplot[color=Blue,mark=*,]
    coordinates {(0.0001,0.0551454624402954)(0.001,0.113256113256113)(0.01,0.424265259984928)(0.1,1.52574286554869)(1,0.137736608940797) 
};\addlegendentry{FNO}; 

\addplot[color=Dandelion,mark=*,]
    coordinates {(0.0001,0.542915038355768)(0.001,0.441441441441441)(0.01,0.023737754333082)(0.1,0.0282436010591349)(1,8.18244059605316) 
};\addlegendentry{GT};
    
    \end{loglogaxis}
    \end{tikzpicture}%
    \caption{Comparison of relative error (Equation \eqref{eqn:l2loss}) for burgers equation with varying $\nu$ (Left) and the relative error ratio comparing the relative errors of FNO and GT against IAE-Net at each $\nu$ (Right). Models are trained with $s=1024$, and tested with $s=1024$ testing data.}%
    \label{fig:viscosity}
\end{figure}

Next, we explore the effects of viscosity on model performance. A data set of 1000 training data and 100 testing data is generated with $\nu=1e^{-4}$ and tested with IAE-Net, FNO, FT, GT, and DeepONet. The results for $\nu=1e^{-4}$ is shown in Figure \ref{fig:burgersinviscid}. IAE-Net achieves the best performance compared to the other baselines. Further experiments are conducted to test the performance of IAE-Net, FNO and GT for varying degrees of $\nu$, using 1000 training and 100 testing data for each $\nu=1e^{-4},1e^{-3},1e^{-2},1e^{-1}$ and $1$. The results comparing the performance of the models across different viscosities and the relative error ratio comparing the errors of FNO and GT against IAE-Net at each $\nu$ are shown in Figure \ref{fig:viscosity}. This ratio is computed using $\frac{m_e-i_e}{i_e}$, where the $L2$ relative error of FNO or GT is denoted by $m_e$ and the $L2$ relative error of IAE-Net is denoted by $i_e$.

IAE-Net outperforms FNO and GT on all viscosities, demonstrating the generalization capability of the model. In contrast, we observe varied performance in FNO and GT. For $\nu=1$, GT fails to achieve good results. On the other viscosities, it can be seen that the performance of FNO starts to deviate from IAE-Net as $\nu$ increases to $0.1$, while the performance of GT deviates from IAE-Net as $\nu$ decreases to $0.0001$. Due to the focus of these methods on well-posed problems, the performance of these baselines starts to deviate when applied to the same problem, but under different well-posedness conditions.

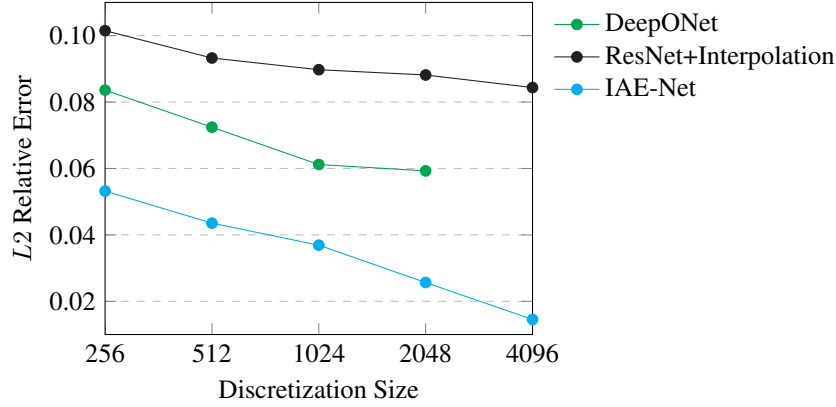
\begin{figure}[t]
    \centering
    \begin{tikzpicture}
    \begin{axis}[
        width=0.44\columnwidth,
        xlabel={Discretization Size},
        ylabel={$L2$ Relative Error},
        xmin=1, xmax=5,
        ymin=0.01, ymax=0.11,
        ytick={0.02,0.04,0.06,0.08,0.10},
        yticklabels={0.02,0.04,0.06,0.08,0.10},
        xtick={1,2,3,4,5},
        xticklabels={256,512,1024,2048,4096},
        legend cell align={left},
        legend style={draw=none},
        legend pos= outer north east,
        ymajorgrids=true,
        grid style=dashed,
    ]
    
\addplot[color=Green,mark=*,]
    coordinates {(1,0.0836)(2,0.07243)(3,0.06119)(4,0.05928) 
};\addlegendentry{DeepONet}; 

\addplot[color=Black,mark=*,]
    coordinates {(1,0.1015)(2,0.09326)(3,0.08974)(4,0.08815)(5,0.08436)
};\addlegendentry{ResNet+Interpolation}; 

\addplot[color=Cyan,mark=*,]
    coordinates {(1,0.0532)(2,0.04354)(3,0.03691)(4,0.02566)(5,0.01454)
};\addlegendentry{IAE-Net};
    
    \end{axis}
    \end{tikzpicture}%
    \caption{$L2$ relative error (Equation \ref{eqn:l2loss}) on the Burgers data set with non-uniform grid. ResNet+Interpolation and IAE-Net are trained with randomly subsampled sorted points of sizes $s=256,512,1024,2048,4096$ and tested on the same resolutions. As DeepONet is not discretization invariant to the inputs, a separate training is performed for each resolution.}
    \label{fig:burgersnonuniform}
\end{figure}

Next, we conduct an experiment to test IAE-Net on inputs with a non-uniform grid. IAE-Net is trained and tested on the Burgers equation data set with viscosity $\nu=1e^{-1}$. The non-uniform grids are obtained via randomly subsampling $s=256,512,1024,2048,4096$ sorted points from the size $s=8192$ data. The subsampling is done such that the end-points $x=0$ and $x=1$ are always included, to ensure that the boundary condition is satisfied. The performance of IAE-Net is compared against DeepONet and ResNet+Interpolation, two baseline models that can handle non-uniform grids. Due to out-of-memory issues when loading the DeepONet data set for larger resolutions, the results for those resolutions are omitted. The results are shown under Figure \ref{fig:burgersnonuniform}. IAE-Net achieves the lowest $L2$ relative error as compared to the other baselines, showing how IAE-Net can also be applied to non-uniform grids.

\subsection{Forward and Inverse Problems}

IAE-Net is evaluated on two Forward and Inverse Problems.

\subsubsection{Darcy Flow}
\label{sec:darcy}

IAE-Net is demonstrated on the forward problem solving the Darcy flow equation. The Darcy flow equation is given by
\begin{equation}\label{eqn:darcy}
    \begin{split}
        -\nabla\cdot (a(x)\nabla u(x))&=f(x),\quad x\in(0,1)^2\\
        u(x)&=0,\quad x\in\partial(0,1)^2,
    \end{split}
\end{equation}
with a Dirichlet boundary condition, with $a$ as the diffusion coefficient and $f$ the forcing function. In this problem, the forward problem is defined as the mapping from the diffusion coefficient $a$ to the solution $u$, i.e., $a(\cdot)\rightarrow u(\cdot)$ for $x\in(0,1)^2$. The Darcy Flow is used as the 2D benchmark problem in \cite{fno,galerkintransformer}, and also explored in works like \cite{bayesian}.

In the benchmark data set, the coefficients $a(x)$ are generated according to $a\sim\mu$ where $\mu=\psi(\mathcal{N}(0,(-\Delta+\tau^2I)^{-\alpha}))$, $\tau=3$ and $\alpha=2$ with
zero Neumann boundary conditions on the Laplacian. The mapping $\psi:\mathbb{R}\rightarrow\mathbb{R}$ takes the value 12 on the positive part of the real line and 3 on the negative, and the forcing function is taken to be $f(x)=1$. The solutions $u$ are obtained using a second-order finite difference scheme. The code to generate the data set is provided by \cite{fno} in \url{https://github.com/zongyi-li/fourier_neural_operator}. A total of 1000 training data and 100 testing data obtained from the benchmark data set is used for the experiment.

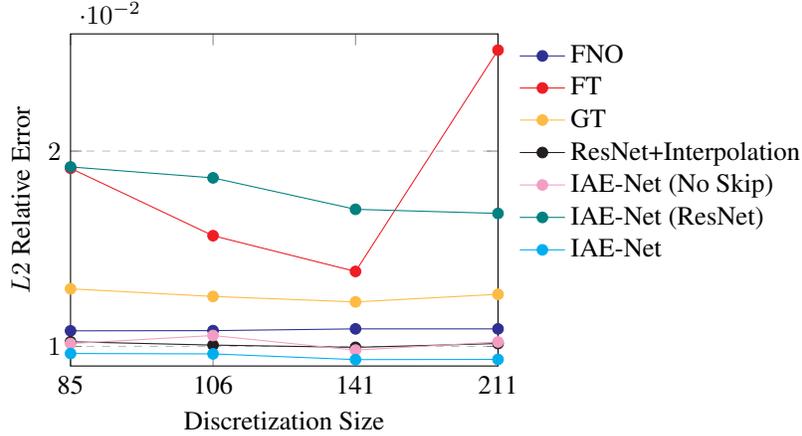
\begin{figure}[t]
    \centering
    \begin{tikzpicture}
    \begin{axis}[
        width=0.44\columnwidth,
        xlabel={Discretization Size},
        ylabel={$L2$ Relative Error},
        xmin=1, xmax=4,
        ymin=0.009, ymax=0.026,
        ytick={0.01,0.02},
        xtick={1,2,3,4},
        xticklabels={85,106,141,211},
        legend cell align={left},
        legend style={draw=none},
        legend pos= outer north east,
        ymajorgrids=true,
        grid style=dashed,
    ]
    
\addplot[color=Blue,mark=*,]
    coordinates {(1,0.0108)(2,0.01081)(3,0.0109)(4,0.0109) 
};\addlegendentry{FNO}; 

\addplot[color=Red,mark=*,]
    coordinates {(1,0.01912)(2,0.01567)(3,0.01384)(4,0.02517) 
};\addlegendentry{FT}; 

\addplot[color=Dandelion,mark=*,]
    coordinates {(1,0.01295)(2,0.01256)(3,0.01228)(4,0.01267) 
};\addlegendentry{GT}; 

\addplot[color=Black,mark=*,]
    coordinates {(1,0.01024)(2,0.01007)(3,0.009954)(4,0.01015) 
};\addlegendentry{ResNet+Interpolation}; 

\addplot[color=Lavender,mark=*,]
    coordinates {(1,0.01015)(2,0.01056)(3,0.009814)(4,0.01023) 
};\addlegendentry{IAE-Net (No Skip)}; 

\addplot[color=teal,mark=*,]
    coordinates {(1,0.01919)(2,0.01863)(3,0.01702)(4,0.01681) 
};\addlegendentry{IAE-Net (ResNet)}; 

\addplot[color=Cyan,mark=*,]
    coordinates {(1,0.009639)(2,0.00962)(3,0.009327)(4,0.009337) 
};\addlegendentry{IAE-Net};
    
    \end{axis}
    \end{tikzpicture}%
    \caption{$L2$ relative error (Equation \eqref{eqn:l2loss}) on the benchmark darcy data set. Models are trained with $s=141$ size training data and tested on the other resolutions.}
    \label{fig:darcybenchmark}
\end{figure}

The models are trained with $s=141$ training data and tested on testing data for each of the resolutions $s=85,106,141,211$. The results are shown in Figure \ref{fig:darcybenchmark}. In this baseline, IAE-Net outperforms all baselines, achieving state-of-the-art error. In this problem, it can be seen that the use of ResNet skip connections in IAE-Net (ResNet) does not perform well. One possible explanation is the difference in the structure of $a$, a piecewise constant function, with $u$, which is smooth, thus the residuals from the ResNet skip connection may not be as important to producing the final output $u$. Surprisingly, it was observed that the performance of the ResNet+Interpolation model performs better than FNO, FT, and GT. This observation suggests that data-driven learning may provide additional benefits to the overall performance, by learning key data-driven features, in this problem. Thus, IAE-Net, which contains the data-driven IAE kernels, is able to outperform existing discretization invariant methods in this application.

% \begin{figure}[t]
%     \centering
%     \includegraphics[width=0.6\columnwidth]{images/darcy.pdf}
%     \caption{$L2$ relative error (Equation \eqref{eqn:l2loss}) on the benchmark darcy data set. Models are trained with $s=141$ size training data and tested on the other resolutions.}
%     \label{fig:darcybenchmark}
% \end{figure}

Next, the performance of IAE-Net is tested on the Darcy problem in a triangular domain with a notch \cite{Lu2022}. This experiment evaluates the performance of IAE-Net when applied to general domains. Following the Darcy flow equation (Equation \ref{eqn:darcy}) with $a(x)=0.1$ and $f=-1$, the operator to be learnt is the mapping from the boundary values $u(x)|_{\partial \Omega}\rightarrow u(x)$, where $\Omega$ is defined as a 2D triangular domain where the corners are given by the following coordinates: $(0,0)$, $(1,0)$ and $(\frac{1}{2},\frac{\sqrt{3}}{2})$, with a rectangular notch where the corners are given by the following coordinates: $(0.49,0)$, $(0.51,0)$, $(0.51,0.4)$ and $(0.49,0.4)$, where $x\in\Omega$. Each boundary of the triangular domain in $\Omega$ is a function $h(y)$, where $y\in[0,1]$, generated via a Gaussian process $\mathcal{GP}$ given by
\begin{equation}\label{eqn:gaussianprocess}
    \begin{split}
        h(y)&\sim\mathcal{GP}(0,\mathcal{K}(y,y')),\\
        \mathcal{K}(y,y')&=\exp[-\frac{(y-y')^2}{2l^2}],\quad l=0.2,\\
        y,y'&\in[0,1],
    \end{split}
\end{equation}
while the boundary values at the boundaries of the notch are fixed as $h(x)=1$. The ground truth is simulated using the PDE Toolbox in Matlab \cite{matlabpde}.

IAE-Net is trained following the data set provided for FNO training by the original paper \cite{Lu2022} in \url{https://github.com/lu-group/deeponet-fno}, which consists of 2000 samples, of which 1900 samples are used as training data and 100 samples are used as test data. IAE-Net is compared against 3 models found in \cite{Lu2022}: DeepONet \cite{chenchen,deeponet}, POD-DeepONet \cite{Lu2022} which replaces the trunk net in DeepONet with a proper orthogonal decomposition (POD) on the training data, and dgFNO+ \cite{Lu2022}, which extends FNO \cite{fno} to a general domain. Due to out-of-memory issues when loading the DeepONet and POD-DeepONet data set for larger resolutions, the results for those resolutions are omitted. The $L2$ relative error (see Equation \ref{eqn:l2loss}) is shown in Table \ref{tab:darcytriangular}. In this application, IAE-Net achieves the lowest $L2$ relative error compared to the other baselines.

\begin{table}[H]
    \centering
    \caption{Results on the Darcy problem in a triangular domain with a notch. IAE-Net is trained with $s=50$ data, and tested on $s=50,100,200$ resolutions. Errors are the $L2$ relative error described in Equation \eqref{eqn:l2loss}, scaled by $1e^2$ to show percentage error. Results labelled with $^*$ are obtained from \cite{Lu2022}.}
    \begin{tabular}{c|ccc}
        \textbf{Model Name} & \textbf{50} & \textbf{100} & \textbf{200} \\
        \hline
        $DeepONet^*$ & 2.64\% \\
        $POD-DeepONet^*$ & 1.00\% \\
        $dgFNO+$ & 7.821\% & 7.819\% & 7.822\% \\
        \hline
        \textbf{IAE-Net} & 0.8094\% & 0.8136\% & 0.8317\% \\
    \end{tabular}
    \label{tab:darcytriangular}
\end{table}

\subsubsection{Scattering Problem}
\label{sec:scattering}

The inhomogeneous media scattering problem \cite{switchnet} with a fixed frequency $\omega$ is modelled by the Helmholtz operator
\begin{equation}\label{eqn:scattering}
    Lu:=(-\Delta-\frac{\omega^2}{c^2(x)})u,
\end{equation}
where $c(x)$ is the velocity field. In most settings, a known background velocity field $c_0(x)$ exists, such that except for a compact domain $\Omega$, $c(x)$ is identical to $c_0(x)$. Thus, by introducing a scatterer $\eta(x)$ that is compactly supported in $\Omega$, i.e.,
\begin{equation}\label{eqn:scatterer}
    \frac{\omega^2}{c(x)^2}=\frac{\omega^2}{c_0(x)^2}+\eta(x),
\end{equation}
then it is possible to work equivalently with $\eta(x)$ instead of $c(x)$. In real-world applications, observation data $d(\cdot)$, derived from the Green's function $G=L^{-1}$ of the Helmholtz operator $L$, is used to evaluate $\eta(\cdot)$. In this field, both the forward map $\eta\rightarrow d$ and the inverse map $d\rightarrow \eta$ are useful for their numerical solutions of the scattering problems. The former provides an alternative to expensive numerical PDE solvers for the Helmholtz equation, while the latter allows one to determine the scatterers from the scattering field, without the usual iterative procedure. The data set is obtained from \cite{switchnet}. 

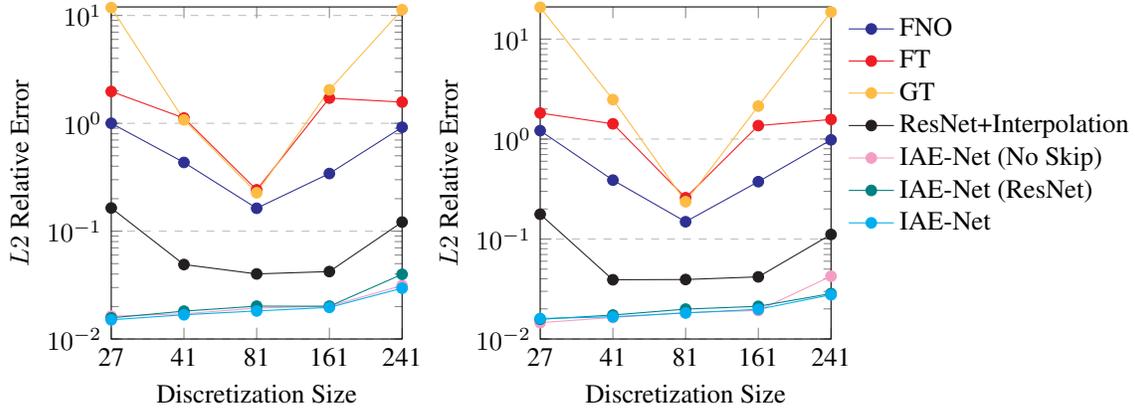
\begin{figure}[t]
    \centering
    \begin{tikzpicture}
    \begin{semilogyaxis}[
        width=0.33\columnwidth,
        xlabel={Discretization Size},
        ylabel={$L2$ Relative Error},
        xmin=1, xmax=5,
        ymin=0.01, ymax=12,
        xtick={1,2,3,4,5},
        xticklabels={27,41,81,161,241},
        legend cell align={left},
        legend style={draw=none},
        legend pos= outer north east,
        ymajorgrids=true,
        grid style=dashed,
    ]
    
\addplot[color=Blue,mark=*,]
    coordinates {(1,1.001)(2,0.4336)(3,0.163)(4,0.3422)(5,0.9192) 
};\addlegendentry{FNO}; 

\addplot[color=Red,mark=*,]
    coordinates {(1,1.974)(2,1.119)(3,0.2416)(4,1.712)(5,1.573) 
};\addlegendentry{FT}; 

\addplot[color=Dandelion,mark=*,]
    coordinates {(1,11.82)(2,1.076)(3,0.227)(4,2.043)(5,11.31) 
};\addlegendentry{GT}; 

\addplot[color=Black,mark=*,]
    coordinates {(1,0.164)(2,0.04903)(3,0.04012)(4,0.0422)(5,0.1214) 
};\addlegendentry{ResNet+Interpolation}; 

\addplot[color=Lavender,mark=*,]
    coordinates {(1,0.01627)(2,0.0171)(3,0.01934)(4,0.02016)(5,0.03131) 
};\addlegendentry{IAE-Net (No Skip)}; 

\addplot[color=teal,mark=*,]
    coordinates {(1,0.0157)(2,0.01812)(3,0.02016)(4,0.02013)(5,0.0398) 
};\addlegendentry{IAE-Net (ResNet)}; 

\addplot[color=Cyan,mark=*,]
    coordinates {(1,0.01506)(2,0.01677)(3,0.0182)(4,0.01963)(5,0.02958) 
};\addlegendentry{IAE-Net};
    
    \legend{}
    \end{semilogyaxis}
    \end{tikzpicture}%
    \begin{tikzpicture}
    \begin{semilogyaxis}[
        width=0.33\columnwidth,
        xlabel={Discretization Size},
        ylabel={$L2$ Relative Error},
        xmin=1, xmax=5,
        ymin=0.01, ymax=21,
        xtick={1,2,3,4,5},
        xticklabels={27,41,81,161,241},
        legend cell align={left},
        legend style={draw=none},
        legend pos= outer north east,
        ymajorgrids=true,
        grid style=dashed,
    ]
    
\addplot[color=Blue,mark=*,]
    coordinates {(1,1.214)(2,0.3885)(3,0.1489)(4,0.3748)(5,0.9825) 
};\addlegendentry{FNO}; 

\addplot[color=Red,mark=*,]
    coordinates {(1,1.821)(2,1.421)(3,0.2588)(4,1.364)(5,1.57) 
};\addlegendentry{FT}; 

\addplot[color=Dandelion,mark=*,]
    coordinates {(1,20.84)(2,2.483)(3,0.2361)(4,2.137)(5,18.66) 
};\addlegendentry{GT}; 

\addplot[color=Black,mark=*,]
    coordinates {(1,0.177)(2,0.03915)(3,0.03936)(4,0.04184)(5,0.1111) 
};\addlegendentry{ResNet+Interpolation}; 

\addplot[color=Lavender,mark=*,]
    coordinates {(1,0.01456)(2,0.01641)(3,0.0184)(4,0.01927)(5,0.04246) 
};\addlegendentry{IAE-Net (No Skip)}; 

\addplot[color=teal,mark=*,]
    coordinates {(1,0.01574)(2,0.01736)(3,0.0199)(4,0.02123)(5,0.02846) 
};\addlegendentry{IAE-Net (ResNet)}; 

\addplot[color=Cyan,mark=*,]
    coordinates {(1,0.01603)(2,0.01666)(3,0.01821)(4,0.01981)(5,0.0277) 
};\addlegendentry{IAE-Net};
    
    \end{semilogyaxis}
    \end{tikzpicture}%

    \caption{$L2$ relative error (Equation \eqref{eqn:l2loss}) on the scattering data set for the forward (Left) and inverse (Right) problem. Model is trained with $s=81$ and tested on different resolutions.}%
    \label{fig:scatteringbenchmark}
\end{figure}

For the scattering problem, the model is trained with 10000 $s=81$ training data and tested on 1000 testing data for each of the resolutions $s=27,41,81,161,241$ for both the forward and inverse problems. The data is generated using $\omega=18\pi$. The results are shown in Figure \ref{fig:scatteringbenchmark}. IAE-Net succeeds in this application while FNO, FT, and GT fail in this case, achieving more than 10 times higher error than IAE-Net. In this application, it was also observed that the performance of ResNet+Interpolation model performs better than FNO, FT, and GT. Similar to the Darcy problem, this suggests the benefits of including data-driven learning via the use of IAE in IAE-Net, which helped to improve the performance of the model, allowing IAE-Net to achieve the best performance.

In addition, all the baseline models show signs of deterioration in resolutions away from the trained resolution $s=81$. Using the proposed data augmentation, this error is alleviated in IAE-Net. In this case, where the problem becomes ill-posed, the performance of models like FNO, FT, and GT which focus on mathematically well-posed problems start to fall off. Including data augmentation in IAE-Net allows the model to capture features on different scales, while the use of data-driven IAE helps to learn suitable features in the data set, thus leading to consistent performance.

\begin{figure}[t]
    \centering
    \begin{tikzpicture}
    \begin{axis}[
        width=0.44\columnwidth,
        xlabel={Discretization Size},
        ylabel={$L2$ Relative Error},
        xmin=1, xmax=3,
        ymin=0.03, ymax=0.22,
        scaled y ticks=base 10:{1},
        xtick={1,2,3},
        xticklabels={81,108,162},
        legend cell align={left},
        legend style={draw=none},
        legend pos= outer north east,
        ymajorgrids=true,
        grid style=dashed,
    ]
    
\addplot[color=orange,mark=*,]
    coordinates {(1,0.2176)(2,0.142)(3,0.1112) 
};\addlegendentry{IAE-Net (Sequential)}; 

\addplot[color=Lavender,mark=*,]
    coordinates {(1,0.07884)(2,0.08372)(3,0.08338) 
};\addlegendentry{IAE-Net (No Skip)}; 

\addplot[color=teal,mark=*,]
    coordinates {(1,0.05022)(2,0.06064)(3,0.05291) 
};\addlegendentry{IAE-Net (ResNet)}; 

\addplot[color=Cyan,mark=*,]
    coordinates {(1,0.04026)(2,0.04773)(3,0.04092) 
};\addlegendentry{IAE-Net};
    
    \end{axis}
    \end{tikzpicture}%
    \caption{$L2$ relative error (Equation \eqref{eqn:l2loss}) on the scattering data set using different data generated with different resolutions. Each resolution consists of 10000 data generated independently from each other, forming a combined total of 30000 data. Sequential training is performend by training the model through loading the data sets in sequence. The last data set in the sequential training is the size $s=162$ data set.}
    \label{fig:scatteringmixed}
\end{figure}
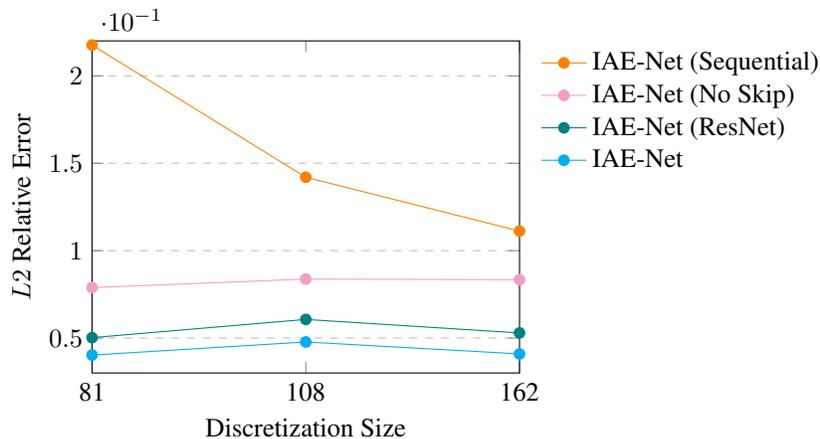

As mentioned in the introduction, aside from the application of IAE-Net to zero-shot learning, an additional experiment is conducted to explore the application of IAE-Net to one more situation: data sets from multiple resolutions are used to perform training. For this experiment, a combined data set is assembled consisting of 10000 data points generated independently for each of the resolutions $s=81,108,162$, forming a combined total of 30000 data. The data is generated using $\omega=18\pi$. Training is done without resizing the data. As existing deep learning software like Pytorch and Tensorflow do not support batching of data with different resolutions in one batch, additional methods to batch the data need to be considered. Here, two different methods of training are considered: 1) Each data set is loaded sequentially and trained for $10$ epochs before loading the next data set. This model is denoted as IAE-Net (Sequential). 2) In each epoch, a random batch of data is loaded from each resolution, and each batch is passed separately into the model. The losses computed for each batch are summed for backpropagation. The results are shown in Figure \ref{fig:scatteringmixed}. We observe that IAE-Net (Sequential) performs poorly, where the model does not converge well, whereas the second method shows better performance, reaching good test error in IAE-Net.

\subsection{Image Processing}
\label{sec:ellipses}

For image processing, IAE-Net is applied to solve a denoising problem. In medical computed tomography (CT) imaging problems, an interesting application is to learn the reconstruction of phantom images from the corresponding filtered backprojection (FBP) CT images with noise added to the projections \cite{ellipses}. In this application, the projections are expressed in terms of the ray transform $\mathcal{P}:\mathcal{X}\rightarrow \mathcal{Y}$, where $\mathcal{X}$ represents the image function space, and $\mathcal{Y}$ is the measurement function space with $\mathcal{P}$ given by
\begin{equation}\label{eqn:ellipses}
    \mathcal{P}(f)(l)=\int_lf(x)dx,\quad l\in\mathbb{M},
\end{equation}
where $l$ are lines from an acquisition geometry $\mathbb{M}$. The inverse operator $\mathcal{P}^{-1}$ is of practical importance, as it allows us to reconstruct the unknown density $f\in\mathcal{X}$ from it's known attenuation data $\mathcal{P}(f)$. 
This inverse problem is ill-posed, that is, given attenuation data $\mathcal{P}(f)$, suppose a solution $f$ to the inverse problem exists, it is unstable with respect to $\mathcal{P}(f)$, where small changes to $\mathcal{P}(f)$ results in large changes to the reconstruction. Thus, the impact of noise added to the projections makes the problem difficult to solve, and solving the problem directly typically leads to over-fitting against the projection data. A solution is to consider a knowledge-driven reconstruction. That is, instead of learning the direct inverse $\mathcal{P}^{-1}$ using an NN, a knowledge-driven component $\mathcal{A}:\mathcal{Y}\rightarrow \mathcal{X}$ is considered and the inverse is learned using
\begin{equation}\label{eqn:ellipses2}
    \mathcal{P}^{-1}=\mathcal{B}_{\theta_b}\circ \mathcal{A}\circ \mathcal{C}_{\theta_c},
\end{equation}
where $\mathcal{A}$ is a known component, while $\mathcal{B}_{\theta_b}:\mathcal{X}\rightarrow \mathcal{X}$ and $\mathcal{C}_{\theta_c}:\mathcal{Y}\rightarrow \mathcal{Y}$ are NNs parameterised by $\theta_b$ and $\theta_c$ respectively. In CT image problems, $\mathcal{A}$ is given by the FBP operator.

A simple data set illustrating this problem is the ellipses data set obtained from \cite{ellipses} via the link \url{https://github.com/adler-j/learned_gradient_tomography}. In this data set, the projection geometry is selected as sparse 30 view parallel beam geometry with 5\% additive Gaussian noise added to the projections. $\mathcal{C}$ is taken as identity, and thus the problem becomes a denoising problem from the FBP CT image with noise back to the original phantom. A particular feature of this inverse problem is that the Ray Transform is used to obtain the data set projections, thus the use of Fourier Transform which is suitable for many of the benchmark problems like the burgers equation and darcy flow in \cite{fno,galerkintransformer} without suitable data-driven learning may not be suitable to this task.

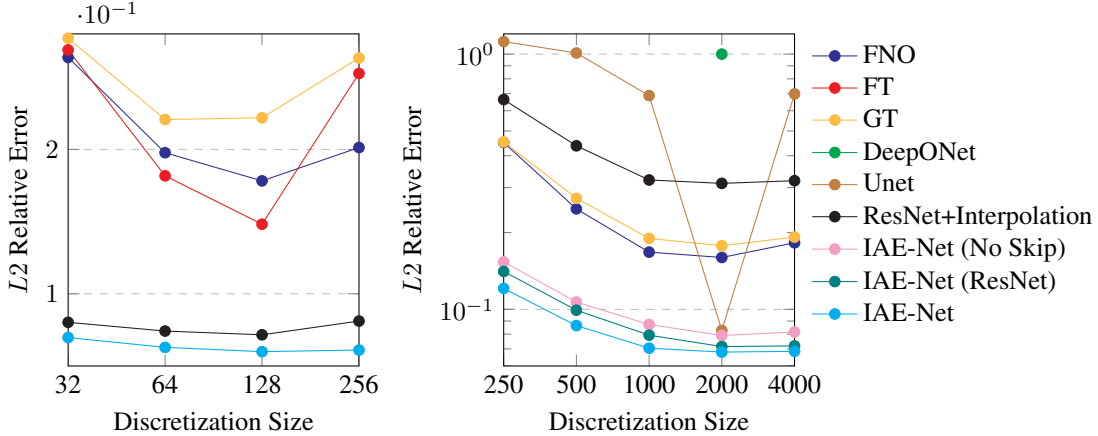
\begin{figure}[t]
    \centering
    \begin{tikzpicture}
     \begin{axis}[
     width=0.33\columnwidth,
     xlabel={Discretization Size},
     ylabel={$L2$ Relative Error},
     xmin=1, xmax=4,
     ymin=0.05, ymax=0.28,
     scaled y ticks=base 10:{1},
     xtick={1,2,3,4},
     xticklabels={32,64,128,256},
     legend cell align={left},
     legend style={draw=none},
     legend pos= outer north east,
     ymajorgrids=true,
     grid style=dashed,
 ]

\addplot[color=Blue,mark=*,]
 coordinates {(1,0.2637)(2,0.1977)(3,0.1782)(4,0.2013) 
};\addlegendentry{FNO}; 

 \addplot[color=Red,mark=*,]
 coordinates {(1,0.2689)(2,0.1817)(3,0.1482)(4,0.2526) 
};\addlegendentry{FT}; 

 \addplot[color=Dandelion,mark=*,]
 coordinates {(1,0.277)(2,0.2207)(3,0.2219)(4,0.2633) 
};\addlegendentry{GT}; 

 \addplot[color=Black,mark=*,]
 coordinates {(1,0.08024)(2,0.07419)(3,0.0716)(4,0.08112) 
};\addlegendentry{ResNet+Interpolation}; 

 \addplot[color=Cyan,mark=*,]
 coordinates {(1,0.06971)(2,0.06289)(3,0.05993)(4,0.06106) 
};\addlegendentry{IAE-Net};

 \legend{}
 \end{axis}
 \end{tikzpicture}
    \begin{tikzpicture}
     \begin{semilogyaxis}[
     width=0.33\columnwidth,
     xlabel={Discretization Size},
     ylabel={$L2$ Relative Error},
     xmin=1, xmax=5,
     ymin=0.06, ymax=1.2,
     % scaled y ticks=base 10:{1},
     % ytick={0.01,0.02},
     xtick={1,2,3,4,5},
     xticklabels={250,500,1000,2000,4000},
     legend cell align={left},
     legend style={draw=none},
     legend pos= outer north east,
     ymajorgrids=true,
     grid style=dashed,
     ]
\addplot[color=Blue,mark=*,]
 coordinates {(1,0.4507)(2,0.2475)(3,0.1676)(4,0.1597)(5,0.1823) 
};\addlegendentry{FNO}; 

\addplot[color=Red,mark=*,]
    coordinates {(1,1000)
};\addlegendentry{FT}; %% PURELY FOR THE LEGEND

\addplot[color=Dandelion,mark=*,]
 coordinates {(1,0.453)(2,0.2724)(3,0.1897)(4,0.1775)(5,0.192) 
};\addlegendentry{GT}; 

\addplot[color=Green,mark=*,]
 coordinates {(4,0.9999) 
};\addlegendentry{DeepONet}; 

\addplot[color=brown,mark=*,]
 coordinates {(1,1.12)(2,1.01)(3,0.6878)(4,0.08274)(5,0.6985) 
};\addlegendentry{Unet}; 

\addplot[color=Black,mark=*,]
 coordinates {(1,0.6637)(2,0.4373)(3,0.3213)(4,0.3116)(5,0.3192) 
};\addlegendentry{ResNet+Interpolation}; 

\addplot[color=Lavender,mark=*,]
 coordinates {(1,0.1536)(2,0.1068)(3,0.08723)(4,0.07904)(5,0.08153) 
};\addlegendentry{IAE-Net (No Skip)}; 

\addplot[color=teal,mark=*,]
 coordinates {(1,0.1408)(2,0.09924)(3,0.07925)(4,0.0715)(5,0.07192) 
};\addlegendentry{IAE-Net (ResNet)}; 

\addplot[color=Cyan,mark=*,]
 coordinates {(1,0.1208)(2,0.08638)(3,0.07048)(4,0.06802)(5,0.06848) 
};\addlegendentry{IAE-Net};
    
     \end{semilogyaxis}
     \end{tikzpicture}
    \caption{$L2$ relative error (Equation \eqref{eqn:l2loss}) on the ellipses data set (Left) and the fecgsyndb data set (Right). The models are trained with s=128 and s=2000 for the two data sets respectively, and tested on different resolutions.}
    \label{fig:ellipsesbenchmark}
    \label{fig:fecgsyndbdiscretizationinvariance}
\end{figure}

For this data set, the model is trained with 10000 $s=128$ training data and tested on 1000 data for each of the resolutions $s=32,64,128,256$, for IAE-Net, FNO, FT, and GT. The choices of the baselines are used to reflect the suitability of using just the Fourier Transform in this task. The results are presented in Figure \ref{fig:ellipsesbenchmark}. IAE-Net achieves the best performance in this task, having about 3 times lower error as compared to the baselines FNO, FT, and GT. This highlights the importance of adopting a learnable data-driven integral kernel for application to general tasks. In this application, the baselines FNO, FT, and GT, which primarily uses Fourier Transform to learn spectral information in the data, may not be suitable to model the denoising problem given by $\mathcal{A}\circ \mathcal{P}$ which uses the Ray Transform. On the other hand, the use of data-driven IAE in IAE-Net can overcome this limitation.

\subsection{Signal Processing}
\label{sec:fecgsyndb}

IAE-Net is also tested on signal processing problems. In this case, the target problem is in signal source separation. In signal source separation, we are given a mixture $f(t)=(f_1(t),\dots,f_m(t))$ for $t\in[0,1]$ obtained from the mixture of a set of $n$ individual source signals, $s(t)=(s_1(t),\dots,s_n(t))$ mixed through a matrix $A$ of size $m\times n$ with some noise function $\eta(t)$. That is, the mixture $f(t)$ is formulated by
\begin{equation}\label{eqn:signal_separation}
    f(t)=A\cdot s(t)+\eta(t).
\end{equation}
The objective is to determine the signals $s(t)$ from the observed mixture $f(t)$. An example of this problem is in the signal separation of maternal and fetal electrocardiogram (ECG) during pregnancy \cite{ecgseparation}. In this problem, $f(t)$ are the signals measured from non-invasive methods, as these methods are safer. The objective is to obtain the desired signals $s(t)$ consisting of the maternal and fetal ECG signals (denoted by mECG and fECG respectively). In addition to the mECG and fECG signals, noise is also present in the mixtures, for example, respiratory noise. The ECG signals are also highly oscillatory, making it difficult for models like FNO and GT, which focuses on well-posed problems, to succeed in this application. Thus, this experiment demonstrates the generalization capability of IAE-Net in ill-posed problems. 

%In practice, existing solutions in signal separation applies autoencoder based models like Unet \cite{unet} and it's variants \cite{scss,waveunet} that are not discretization invariant. The importance of including discretization invariance arises where different events can be captured using different time scales. This applies to many other 1D signal problems like audio signals, making it a good choice of application to test IAE-Net on.

For this experiment, the data set used are ECG mixtures obtained from \cite{fecgsyndb} via the link \url{https://physionet.org/content/fecgsyndb/1.0.0}. The models are trained using 25000 $s=2000$ length signals, and tested on 6500 testing data for each the resolutions $s=250,500,1000,2000,4000$. The results are displayed in Figure \ref{fig:fecgsyndbdiscretizationinvariance}.

Again, IAE-Net succeeds compared to the baselines, which mostly fail to achieve reasonable errors in this application. In this problem, the higher frequency and oscillatory ECG signals suggest that learning in the frequency domain of the Fourier transform can better capture these oscillatory structures. This is not present in baselines like FNO, which selects a fixed number of predetermined modes from the Fourier transform to achieve discretization invariance, or in GT, which learns in the original input domain via the attention blocks, due to their focus on well-posed problems. In IAE-Net, multi-channel learning provides better generalization, through facilitating learning over different transformed domains of the input. In each of these domains, the data-driven integral kernels in IAE promotes the learning of key data-driven features. Even compared to Unet, a popular model for signal processing tasks, IAE-Net achieves lower error, with the added benefit of discretization invariance, which is not present in Unet.% This lack of discretization invariance in Unet thus requires expensive re-training to accommodate different resolutions.

\section{Conclusion}\label{sec:con}

This paper proposes a novel deep learning framework based on integral autoencoders (IAE-Net) for discretization invariant learning. This is achieved via a proposed data-driven IAE for encoding and decoding of inputs of arbitrary discretization to a fixed size, performing data-driven compression and discretization invariant learning. This basic building block is applied in parallel to form wide and multi-channel structures, which are repeatedly composed to form a deep and densely connected neural network with skip connections as IAE-Net. IAE-Net is trained with randomized data augmentation that generates training data with heterogeneous structures to facilitate the performance of discretization invariant learning. Numerical experiments demonstrate powerful generalization performance in numerous applications ranging from predictive data science, solving forward and inverse problems in scientific computing, and signal/image processing. In contrast to existing alternatives in literature, either IAE-Net achieves the best accuracy or existing methods fail to provide meaningful results. The code is available on \url{https://github.com/IAE-Net/iae_net}.

% Acknowledgements should go at the end, before appendices and references

\acks{H.~Yang was partially supported by the US National Science Foundation under award DMS-1945029 and DMS-2206333.}


\bibliographystyle{unsrt}  
\bibliography{sample}

\newpage

\appendix
\section*{Appendices}

%\section{Table of Notations}

\section{Ablation Study}
\label{apd:ablation_study}

This section presents numerical results for an initial ablation study of the main components in IAE-Net. Unless otherwise stated, the data set used is the burgers benchmark data set consisting of 1000 training samples and 100 testing samples provided by \cite{fno} in \url{https://github.com/zongyi-li/fourier_neural_operator}.

\subsection{Effect of Number of Blocks on Performance}
\label{apd:number_of_blocks}

\begin{figure}[H]
    \centering
    \begin{tikzpicture}
    \begin{axis}[
        width=0.44\columnwidth,
        xlabel={Number of $\mathcal{IAE}$ Blocks},
        ylabel={$L2$ Relative Error},
        xmin=1, xmax=5,
        ymin=0.0018, ymax=0.006,
        xtick={1,2,3,4,5},
        xticklabels={1,2,3,4,5},
        legend cell align={left},
        legend style={draw=none},
        legend pos= outer north east,
        ymajorgrids=true,
        grid style=dashed,
    ]

\addplot[color=Cyan,mark=*,]
    coordinates {(1,0.005253)(2,0.004355)(3,0.003128)(4,0.001923)(5,0.0023) 
};\addlegendentry{IAE-Net};
    
    \legend{}
    \end{axis}
    \end{tikzpicture}%
    \caption{$L2$ relative error of IAE-Net with different number of blocks on the burgers data set with $\nu=1e^{-1}$. IAE-Net is trained and tested on the $s=1024$ resolution data.}
    \label{fig:numblocks}
\end{figure}
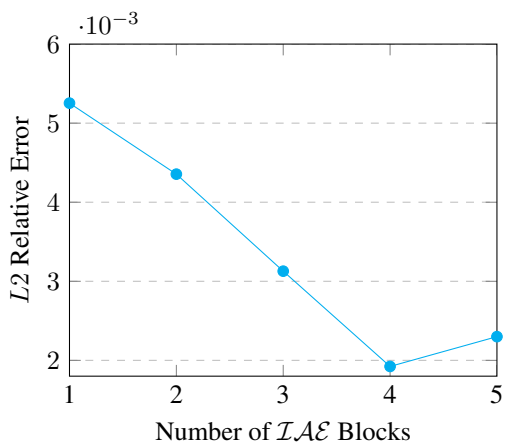

In this section, the number of blocks in IAE-Net is considered for the burgers benchmark data set. IAE-Net is trained with $L=1,\dots,5$ $\mathcal{IAE}$ blocks using the training data of size $s=1024$ and tested on the same resolution, and the relative errors are reported in Figure \ref{fig:numblocks}. It can be seen that incorporating a repeatedly composed structure in IAE-Net improves the accuracy of the model. We note a possible increase in error for $L=5$, suggesting a possibility of overfitting to the burgers data set.

Next, an ablation study is conducted to motivate the choice of a multi-layer recursive structure with 1- or 2-layer NNs for the design of each component in IAE-Net. To do so, we compare the results for IAE-Net with 4 $\mathcal{IAE}$ Blocks designed following Section \ref{sec:training_details} with an IAE-Net with 1 $\mathcal{IAE}$ Block but $\phi_0$, $\phi_1$ and $\phi_2$ (see Section \ref{sec:iae}) are deeper NNs. Instead of a single hidden layer FNN (see Equation \ref{eqn:fnn}), a 4 hidden layer FNN is used for this variation of IAE-Net, which we denote as IAE-Net (Deep). The results on the Burgers benchmark data set \cite{fno} are shown in Table \ref{tab:burgersdeep}. The use of multiple blocks is beneficial to the overall performance of the model, reducing error by 2 times. The multi-layer structure incorporates multiple encoders, allowing each of them to focus on specific features of the problem. Together with the use of skip connections to collate previous approximations, subsequent blocks can make use of previous approximations to learn an encoder which yields a better feature representation of the target.

\begin{table}[H]
    \centering
    \caption{Results on the benchmark Burgers data set comparing performance of IAE-Net and IAE-Net (Deep). Errors are the $L2$ relative error described in Equation \eqref{eqn:l2loss}, scaled by $1e^2$ to show percentage error.}
    \begin{tabular}{c|cccccc}
        \textbf{Model Name} & \textbf{256} & \textbf{512} & \textbf{1024} & \textbf{2048} & \textbf{4096} & \textbf{8192} \\
        \hline
        \textbf{IAE-Net (Deep)} & 0.4137\% & 0.4138\% & 0.4138\% & 0.4139\% & 0.4139\% & 0.4139\% \\
        \textbf{IAE-Net} & 0.1923\% & 0.1924\% & 0.1923\% & 0.1923\% & 0.1924\% & 0.1924\% \\
    \end{tabular}
    \label{tab:burgersdeep}
\end{table}

\subsection{Effect of Multi-Channel Learning on Performance}
\label{apd:comparison_parallel_blocks}

To show the benefits of adopting multiple branches, IAE-Net is tested under two additional variations for the burgers benchmark data set. First, IAE-Net is trained with only the original input as a single IAE branch, denoted by IAE-Net (Original). The other variation considers IAE-Net with only the Fourier transformed IAE branch, denoted by IAE-Net (Fourier). The results are shown in Table \ref{tab:burgersparallel}. Comparing between IAE-Net (Original) and IAE-Net (Fourier), the use of Fourier transform is slightly beneficial and more suitable to the burgers equation as opposed to using the original input. However, when both the branches are used in a 2-Channel framework (see IAE-Net), the performance has improved by more than 2 times.

\begin{table}[H]
    \centering
    \caption{Results on the benchmark burgers data set comparing performance with and without parallel blocks. Errors are the $L2$ relative error described in Equation \eqref{eqn:l2loss}, scaled by $1e^2$ to show percentage error.}
    \begin{tabular}{c|cccccc}
        \textbf{Model Name} & \textbf{256} & \textbf{512} & \textbf{1024} & \textbf{2048} & \textbf{4096} & \textbf{8192} \\
        \hline
        \textbf{IAE-Net (Original)} & 0.5276\% & 0.5275\% & 0.5276\% & 0.5276\% & 0.5275\% & 0.5275\% \\
        \textbf{IAE-Net (Fourier)} & 0.4183\% & 0.4183\% & 0.4184\% & 0.4184\% & 0.4185\% & 0.4184\% \\
        \hline
        \textbf{IAE-Net} & 0.1923\% & 0.1924\% & 0.1923\% & 0.1923\% & 0.1924\% & 0.1924\% \\
    \end{tabular}
    \label{tab:burgersparallel}
\end{table}

\subsection{Effect of Channel MLP for Post Processing in IAE}
\label{apd:transformer_architecture}

Using the burgers benchmark data set, IAE-Net is tested with and without the channel MLP for post processing of each IAE component in $\mathcal{IAE}$. The model without MLP is denoted as IAE-Net (No MLP). The results are shown under Table \ref{tab:transformerarchitecture}. The use of MLP is highly beneficial to the overall performance of the model, reducing error by almost 7 times.

\begin{table}[H]
    \centering
    \caption{Results on the benchmark burgers data set comparing performance with and without channel MLP. Errors are the $L2$ relative error described in Equation \eqref{eqn:l2loss}, scaled by $1e^2$ to show percentage error.}
    \begin{tabular}{c|cccccc}
        \textbf{Model Name} & \textbf{256} & \textbf{512} & \textbf{1024} & \textbf{2048} & \textbf{4096} & \textbf{8192} \\
        \hline
        \textbf{IAE-Net (No MLP)} & 1.491\% & 1.47\% & 1.476\% & 1.517\% & 1.593\% & 1.813\% \\
        \textbf{IAE-Net} & 0.2094\% & 0.2083\% & 0.2082\% & 0.2087\% & 0.2093\% & 0.2096\% \\
    \end{tabular}
    \label{tab:transformerarchitecture}
\end{table}

\subsection{Effect of Non-linear Integral Transform in Integral Kernels}

Using the burgers benchmark data set, IAE-Net is tested with and without the auxiliary information for the nonlinear integral kernels $\phi_1$ and $\phi_2$ in IAE (see Section \ref{sec:iae}). The model without auxiliary information is denoted as IAE-Net (Linear). The results are shown under Table \ref{tab:linear}. The use of auxiliary information reduced the error by almost 2 times.

\begin{table}[H]
    \centering
    \caption{Results on the benchmark burgers data set comparing performance with and without auxiliary information in $\phi_1$ and $\phi_2$ in IAE (see Section \ref{sec:iae}). Errors are the $L2$ relative error described in Equation \eqref{eqn:l2loss}, scaled by $1e^2$ to show percentage error.}
    \begin{tabular}{c|cccccc}
        \textbf{Model Name} & \textbf{256} & \textbf{512} & \textbf{1024} & \textbf{2048} & \textbf{4096} & \textbf{8192} \\
        \hline
        \textbf{IAE-Net (Linear)} & 0.4181\% & 0.4182\% & 0.4183\% & 0.4184\% & 0.4185\% & 0.4185\% \\
        \textbf{IAE-Net} & 0.2094\% & 0.2083\% & 0.2082\% & 0.2087\% & 0.2093\% & 0.2096\% \\
    \end{tabular}
    \label{tab:linear}
\end{table}

\subsection{Effect of Post Processing Blocks in IAE-Net}

Using the Burgers benchmark data set, IAE-Net is tested with and without the post processing block $G$ (see Section \ref{sec:iaenet}). The model without $G$ is denoted as IAE-Net (w/o $G$). The results are shown under Table \ref{tab:postpro}. The post processing blocks reduced the error by almost 4 times, and we observe that the $L2$ relative error (Equation \ref{eqn:l2loss}) across the resolutions is more consistent.

\begin{table}[H]
    \centering
    \caption{Results on the benchmark Burgers data set comparing performance with and without $G$ in IAE-Net (see Section \ref{sec:iaenet}). Errors are the $L2$ relative error described in Equation \eqref{eqn:l2loss}, scaled by $1e^2$ to show percentage error.}
    \begin{tabular}{c|cccccc}
        \textbf{Model Name} & \textbf{256} & \textbf{512} & \textbf{1024} & \textbf{2048} & \textbf{4096} & \textbf{8192} \\
        \hline
        \textbf{IAE-Net (w/o $G$)} & 0.8657\% & 0.8329\% & 0.8446\% & 0.8659\% & 0.8980\% & 0.9520\% \\
        \textbf{IAE-Net} & 0.2094\% & 0.2083\% & 0.2082\% & 0.2087\% & 0.2093\% & 0.2096\% \\
    \end{tabular}
    \label{tab:postpro}
\end{table}

\subsection{Effect of Data Augmentation in Discretization Invariant Models}
\label{apd:data_augment}

\begin{figure}[H]
    \centering
    \begin{tikzpicture}
    \begin{semilogyaxis}[
        width=0.33\columnwidth,
        xlabel={Discretization Size},
        ylabel={$L2$ Relative Error},
        xmin=1, xmax=5,
        ymin=0.01, ymax=2,
        xtick={1,2,3,4,5},
        xticklabels={27,41,81,161,241},
        legend cell align={left},
        legend style={draw=none},
        legend pos= outer north east,
        ymajorgrids=true,
        grid style=dashed,
    ]
    
\addplot[color=Blue,mark=*,]
    coordinates {(1,1.001)(2,0.4336)(3,0.163)(4,0.3422)(5,0.9192) 
};\addlegendentry{FNO};

\addplot[color=Green,mark=*,]
    coordinates {(1,0.3162)(2,0.2308)(3,0.1622)(4,0.245)(5,0.2936) 
};\addlegendentry{FNO (With DA)}; 

\addplot[color=Black,mark=*,]
    coordinates {(1,0.164)(2,0.04903)(3,0.04012)(4,0.0422)(5,0.1214) 
};\addlegendentry{ResNet+Interpolation}; 

\addplot[color=Magenta,mark=*,]
    coordinates {(1,0.06192)(2,0.04487)(3,0.039)(4,0.04364)(5,0.0597) 
};\addlegendentry{ResNet+Interpolation\\(With DA)};

\addplot[color=Cyan,mark=*,]
    coordinates {(1,0.01506)(2,0.01677)(3,0.0182)(4,0.01963)(5,0.02958) 
};\addlegendentry{IAE-Net};
    
    \legend{}
    \end{semilogyaxis}
    \end{tikzpicture}%
    \begin{tikzpicture}
    \begin{semilogyaxis}[
        width=0.33\columnwidth,
        xlabel={Discretization Size},
        ylabel={$L2$ Relative Error},
        xmin=1, xmax=5,
        ymin=0.01, ymax=2,
        xtick={1,2,3,4,5},
        xticklabels={27,41,81,161,241},
        legend cell align={left},
        legend style={draw=none,cells={align=left}},
        legend pos= outer north east,
        ymajorgrids=true,
        grid style=dashed,
    ]
    
\addplot[color=Blue,mark=*,]
    coordinates {(1,1.214)(2,0.3885)(3,0.1489)(4,0.3748)(5,0.9825) 
};\addlegendentry{FNO};

\addplot[color=Green,mark=*,]
    coordinates {(1,0.2845)(2,0.2279)(3,0.1724)(4,0.2396)(5,0.273) 
};\addlegendentry{FNO (With DA)}; 

\addplot[color=Black,mark=*,]
    coordinates {(1,0.177)(2,0.03915)(3,0.03936)(4,0.04184)(5,0.1111) 
};\addlegendentry{ResNet+Interpolation}; 

\addplot[color=Magenta,mark=*,]
    coordinates {(1,0.05822)(2,0.04047)(3,0.03834)(4,0.0406)(5,0.05585) 
};\addlegendentry{ResNet+Interpolation\\(With DA)};

\addplot[color=Cyan,mark=*,]
    coordinates {(1,0.01603)(2,0.01666)(3,0.01821)(4,0.01981)(5,0.0277) 
};\addlegendentry{IAE-Net};
    
    \end{semilogyaxis}
    \end{tikzpicture}%

    \caption{$L2$ relative error (Equation \ref{eqn:l2loss}) on the scattering data set comparing the impact of data augmentation for the forward (Left) and inverse (Right) problem. Model is trained with $s=81$ and tested on different resolutions.}%
    \label{fig:scatteringda}
\end{figure}
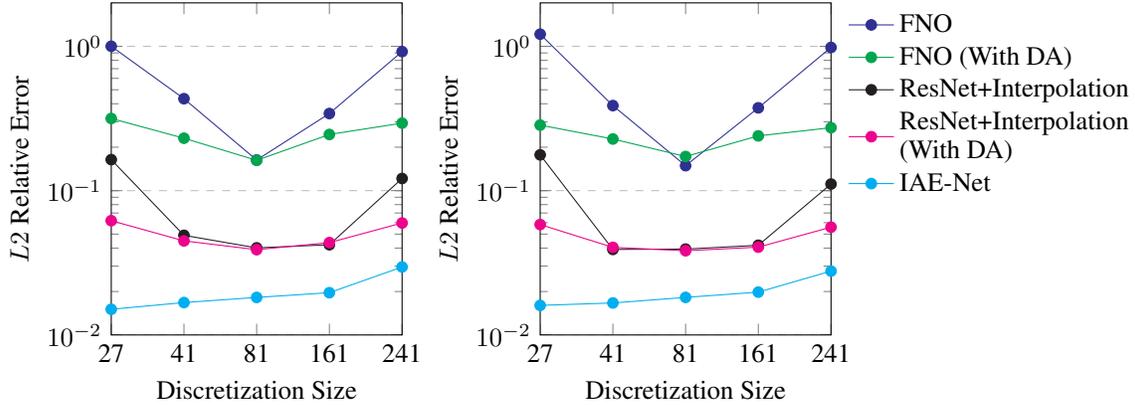

This section shows the effect of the proposed data augmentation (see Section \ref{sec:data_augment}) on two other discretization invariant models, FNO and ResNet+Interpolation, on the Forward and Inverse Scattering Problem (see Section \ref{sec:scattering}). The results for each model with data augmentation are denoted by FNO (With DA) and ResNet+Interpolation (With DA) respectively. Figure \ref{fig:scatteringda} shows the $L2$ relative error (Equation \ref{eqn:l2loss}). The results show that the use of data augmentation helps discretization invariant models to capture features from different scales, reducing the amount of deterioration in resolutions away from the trained resolution.

\section{Comparison of Model Evaluation Speed}
\label{apd:comparison_performance}

%The performance of IAE-Net is compared against the baseline models and the evaluation times are reported. The testing time is computed as the average time taken over 10 trials in seconds to test each model with 100 testing samples from the Burgers benchmark data set with $s=1024$. The results are shown in Table \ref{tab:modelevaluationtimes}. %In addition to the baselines, the time taken to generate each data (using the split step method) numerically is also reported.

The performance of IAE-Net is compared against the baseline models and the evaluation times are reported. The training time is computed as the average time taken over 10 trials in seconds to train each model on 1000 training samples from the Burgers benchmark data set with $s=1024$ for 1 epoch. The testing time is computed as the average time taken over 10 trials in seconds to test each model with 100 testing samples from the Burgers benchmark data set with $s=1024$. The results are shown in Table \ref{tab:modelevaluationtimes}.

% \begin{table}[H]
%     \centering
%     \caption{Comparison for model evaluation times, in seconds, for the different models with the Burgers equation Benchmark data set.}
%     \begin{tabular}{c|ccc}
%         \textbf{Model Name} & \textbf{Test Time (sec)} \\
%         % \hline
%         % Split Step Method & 0.8025 \\
%         \hline
%         $FNO$ & 0.02333 \\
%         $FT$ & 0.1757 \\
%         $GT$ & 0.1687 \\
%         $DeepONet$ & 2.109 \\
%         $Unet$ & 0.424 \\
%         $ResNet+Interpolation$ & 0.05865 \\
%         \hline
%         \textbf{IAE-Net (No DA)} & 0.2725 \\
%         \textbf{IAE-Net (No Skip)} & 0.2866 \\
%         \textbf{IAE-Net (ResNet)} & 0.2717 \\
%         \hline
%         \textbf{IAE-Net} & 0.2715 \\
%     \end{tabular}
%     \label{tab:modelevaluationtimes}
% \end{table}

\begin{table}[H]
    \centering
    \caption{Comparison for model evaluation times, in seconds, for the different models with the Burgers equation benchmark data set.}
    \begin{tabular}{c|ccc}
        \textbf{Model Name} & \textbf{Parameters} & \textbf{Train Time (sec)} & \textbf{Test Time (sec)} \\
        \hline
        $FNO$ & 0.5M & 0.6946 & 0.02333 \\
        $FT$ & 0.5M & 9.311 & 0.1757 \\
        $GT$ & 0.5M & 8.631 & 0.1687 \\
        $DeepONet$ & 4.8M & 21.53 & 2.109 \\
        $Unet$ & 9.2M & 6.741 & 0.424 \\
        $ResNet+Interpolation$ & 5.3M & 6.576 & 0.05865 \\
        \hline
        \textbf{IAE-Net (No DA)} & 5.3M & 5.837 & 0.2725 \\
        \textbf{IAE-Net (No Skip)} & 5.3M & 66.18 & 0.2866 \\
        \textbf{IAE-Net (ResNet)} & 5.3M & 75.61 & 0.2717 \\
        \hline
        \textbf{IAE-Net} & 5.3M & 77.2 & 0.2715 \\
    \end{tabular}
    \label{tab:modelevaluationtimes}
\end{table}

%Of particular interest is the advantage of NN models compared to solving the problem numerically. The use of IAE-Net allows for an evaluation time of almost 3 times faster than the numerical method in solving the Burgers equation. This becomes especially prominent in solving the problem for higher/lower viscosities, where a finer step size is needed for numerical solutions. 
The use of data augmentation (see Section \ref{sec:data_augment}) leads to a longer training time required for IAE-Net. During each iteration, in addition to the training data pair $(\bar{f},\bar{g})$, additional interpolations of the data are used to augment the objective function (see Equation \ref{eqn:loss}). This requires multiple forward and back-propagation passes to train IAE-Net. Also, IAE-Net requires the assembly of a kernel matrix $\bar{K}$ for nonlinear integral transform (see Equation \ref{eqn:numintegraltransform2}), which is expensive for large resolutions. Note, that without data augmentation, IAE-Net (No DA) achieves a faster training time than all the baselines, except FNO.

The fast evaluation time of FNO is due to the assumption of the inputs having uniform grid, allowing for the use of the built in Fourier transform. IAE-Net is able to handle inputs of non-uniform grid, as the pointwise coordinate values are used in the integral kernels (see Section \ref{sec:iae}). In the case of a non-uniform grid, a non-uniform Fourier transform is needed in FNO, which will significantly slow down evaluation time.
It is an interesting future project to explore hierarchical low-rank structures \cite{fan2019solving,Fan2019,switchnet,Li2020} in the matrix-vector multiplications in IAE-Net to speed up the application of IAE-Net.

\section{Tables Used for the Figures in the Experiments}

If there is no result for a certain method in an example, it means either the memory cost is not affordable or the accuracy is too poor. 

\subsection{Burgers Equation}

This section shows the results used to create the figures in Section \ref{sec:burgers}.

\begin{table}[H]
    \centering
    \caption{Results on the burgers data set, rescaled from the actual $L2$ relative error computed using Equation \eqref{eqn:l2loss} by multiplying $1e^{3}$. Models are trained with $s=1024$ and tested on the other resolutions.}
    \begin{tabular}{c|cccccc}
        \textbf{Model Name} & \textbf{256} & \textbf{512} & \textbf{1024} & \textbf{2048} & \textbf{4096} & \textbf{8192} \\
        \hline
        $FNO$ & 0.7597 & 0.7311 & 0.7245 & 0.7232 & 0.7232 & 0.7232 \\
        $FT$ & 0.8104 & 0.8002 & 0.7983 & 0.7974 & 0.7971 & 0.7969 \\
        $GT$ & 0.7992 & 0.7991 & 0.799 & 0.799 & 0.799 & 0.799 \\
        $DeepONet^{\dag}$ & 8.695 & 7.797 & 8.13 & 8.265 & & \\
        $ResNet+Interpolation$ & 48.31 & 47.28 & 46.9 & 46.74 & 46.67 & 46.63 \\
        \hline
        \textbf{IAE-Net (No DA)} & 9.813 & 2.244 & 0.2821 & 5.006 & 23.2 & 35.67 \\
        \hline
        \textbf{IAE-Net (No Skip)} & 0.6326 & 0.6143 & 0.6119 & 0.6108 & 0.6102 & 0.6094 \\
        \textbf{IAE-Net (ResNet)} & 0.4664 & 0.4661 & 0.4663 & 0.4666 & 0.4663 & 0.4671 \\
        \hline
        \textbf{IAE-Net} & 0.2758 & 0.2754 & 0.2753 & 0.2752 & 0.2752 & 0.2751 \\
    \end{tabular}
    \label{tab:burgersbenchmark}
\end{table}

\begin{table}[H]
    \centering
    \caption{Results on the burgers data set with $\nu=1e^{-4}$. Models are trained with $s=1024$ and tested on the other resolutions. Errors are the $L2$ relative error described in Equation \eqref{eqn:l2loss}, scaled by $1e^2$ to show percentage error.}
    \begin{tabular}{c|cccccc}
        \textbf{Model Name} & \textbf{256} & \textbf{512} & \textbf{1024} & \textbf{2048} & \textbf{4096} & \textbf{8192} \\
        \hline
        $FNO$ & 6.885\% & 7.112\% & 7.29\% & 7.259\% & 7.258\% & 7.257\% \\
        $FT$ & 8.801\% & 9.543\% & 9.574\% & 9.562\% & 9.549\% & 9.555\% \\
        $GT$ & 9.791\% & 10.58\% & 10.66\% & 10.65\% & 10.65\% & 10.65\% \\
        $DeepONet^{\dag}$ & 40.07\% & 39.07\% & 38.89\% & 38.13\% & & \\
        $ResNet+Interpolation$ & 31.15\% & 30.97\% & 30.86\% & 30.81\% & 30.79\% & 30.78\% \\
        \hline
        \textbf{IAE-Net (No Skip)} & 8.787\% & 9.37\% & 9.155\% & 9.121\% & 9.079\% & 9.043\% \\
        \textbf{IAE-Net (ResNet)} & 7.325\% & 7.682\% & 7.774\% & 7.791\% & 7.772\% & 7.737\% \\
        \hline
        \textbf{IAE-Net} & 6.599\% & 6.826\% & 6.909\% & 6.91\% & 6.914\% & 6.918\% \\
    \end{tabular}
    \label{tab:burgersinviscid}
\end{table}

The following Tables \ref{tab:burgers001}, \ref{tab:burgers01}, \ref{tab:burgers1} and \ref{tab:burgers10} provides additional results for the other resolutions and the results used to obtain Figures \ref{fig:viscosity} and \ref{fig:viscosity}.

\begin{table}[H]
    \centering
    \caption{Results on the burgers data set with $\nu=1e^{-3}$. Models are trained with $s=1024$ and tested on the other resolutions. Errors are the $L2$ relative error described in Equation \eqref{eqn:l2loss}, scaled by $1e^2$ to show percentage error.}
    \begin{tabular}{c|cccccc}
        \textbf{Model Name} & \textbf{256} & \textbf{512} & \textbf{1024} & \textbf{2048} & \textbf{4096} & \textbf{8192} \\
        \hline
        $FNO$ & 1.746\% & 1.737\% & 1.733\% & 1.732\% & 1.731\% & 1.731\% \\
        $GT$ & 2.239\% & 2.24\% & 2.24\% & 2.241\% & 2.241\% & 2.241\% \\
        \hline
        \textbf{IAE-Net} & 1.561\% & 1.555\% & 1.554\% & 1.555\% & 1.558\% & 1.562\% \\
    \end{tabular}
    \label{tab:burgers001}
\end{table}

\begin{table}[H]
    \centering
    \caption{Results on the burgers data set with $\nu=1e^{-2}$. Models are trained with $s=1024$ and tested on the other resolutions. Errors are the $L2$ relative error described in Equation \eqref{eqn:l2loss}, scaled by $1e^2$ to show percentage error.}
    \begin{tabular}{c|cccccc}
        \textbf{Model Name} & \textbf{256} & \textbf{512} & \textbf{1024} & \textbf{2048} & \textbf{4096} & \textbf{8192} \\
        \hline
        $FNO$ & 0.3879\% & 0.3804\% & 0.378\% & 0.3772\% & 0.3769\% & 0.3767\% \\
        $GT$ & 0.2716\% & 0.2716\% & 0.2717\% & 0.2717\% & 0.2717\% & 0.2717\% \\
        \hline
        \textbf{IAE-Net} & 0.2654\% & 0.2638\% & 0.2622\% & 0.2609\% & 0.2591\% & 0.258\% \\
    \end{tabular}
    \label{tab:burgers01}
\end{table}

\begin{table}[H]
    \centering
    \caption{Results on the burgers data set with $\nu=1e^{-1}$. Models are trained with $s=1024$ and tested on the other resolutions. Errors are the $L2$ relative error described in Equation \eqref{eqn:l2loss}, scaled by $1e^2$ to show percentage error.}
    \begin{tabular}{c|cccccc}
        \textbf{Model Name} & \textbf{256} & \textbf{512} & \textbf{1024} & \textbf{2048} & \textbf{4096} & \textbf{8192} \\
        \hline
        $FNO$ & 0.1674\% & 0.1702\% & 0.1717\% & 0.1725\% & 0.1729\% & 0.1731\% \\
        $GT$ & 0.06992\% & 0.06991\% & 0.0699\% & 0.0699\% & 0.0698\% & 0.0697\% \\
        \hline
        \textbf{IAE-Net} & 0.06975\% & 0.06869\% & 0.06798\% & 0.06696\% & 0.06612\% & 0.06571\% \\
    \end{tabular}
    \label{tab:burgers1}
\end{table}

\begin{table}[H]
    \centering
    \caption{Results on the burgers data set with $\nu=1$. Models are trained with $s=1024$ and tested on the other resolutions. Errors are the $L2$ relative error described in Equation \eqref{eqn:l2loss}, scaled by $1e^2$ to show percentage error.}
    \begin{tabular}{c|cccccc}
        \textbf{Model Name} & \textbf{256} & \textbf{512} & \textbf{1024} & \textbf{2048} & \textbf{4096} & \textbf{8192} \\
        \hline
        $FNO$ & 2.864\% & 2.835\% & 2.825\% & 2.82\% & 2.819\% & 2.818\% \\
        $GT$ & 22.8\% & 22.8\% & 22.8\% & 22.8\% & 22.8\% & 22.8\% \\
        \hline
        \textbf{IAE-Net} & 2.538\% & 2.497\% & 2.483\% & 2.478\% & 2.476\% & 2.474\% \\
    \end{tabular}
    \label{tab:burgers10}
\end{table}

Table \ref{tab:burgersnonuniform} shows the results used to obtain Figure \ref{fig:burgersnonuniform}.

\begin{table}[H]
    \centering
    \caption{Results on the Burgers data set with non-uniform grids. ResNet+Interpolation and IAE-Net is trained with randomly subsampled sorted points of sizes $s=256,512,1024,2048,4096$ and tested on the same resolutions. As DeepONet is not discretization invariant to the inputs, a separate training is performed for each resolution. Errors are the $L2$ relative error described in Equation \eqref{eqn:l2loss}, scaled by $1e^2$ to show percentage error.}
    \begin{tabular}{c|ccccc}
        \textbf{Model Name} & \textbf{256} & \textbf{512} & \textbf{1024} & \textbf{2048} & \textbf{4096} \\
        \hline
        $DeepONet$ & 8.36\% & 7.243\% & 6.119\% & 5.928\% & \\
        $ResNet+Interpolation$ & 10.15\% & 9.326\% & 8.974\% & 8.815\% & 8.436\% \\
        \hline
        \textbf{IAE-Net} & 5.32\% & 4.354\% & 3.691\% & 2.566\% & 1.454\% \\
    \end{tabular}
    \label{tab:burgersnonuniform}
\end{table}

\subsection{Darcy Flow}

This section shows the results used to create the figures in Section \ref{sec:darcy}.

\begin{table}[H]
    \centering
    \caption{Results on the benchmark darcy data set. Models are trained with $s=141$ size training data and tested on the other resolutions. Errors are the $L2$ relative error described in Equation \eqref{eqn:l2loss}, scaled by $1e^2$ to show percentage error.}
    \begin{tabular}{c|cccc}
        \textbf{Model Name} & \textbf{85} & \textbf{106} & \textbf{141} & \textbf{211} \\
        \hline
        $FNO$ & 1.08\% & 1.081\% & 1.09\% & 1.09\% \\
        $FT$ & 1.912\% & 1.567\% & 1.384\% & 2.517\% \\
        $GT$ & 1.295\% & 1.256\% & 1.228\% & 1.267\% \\
        $ResNet+Interpolation$ & 1.024\% & 1.007\% & 0.9954\% & 1.015\% \\
        \hline
        \textbf{IAE-Net (No Skip)} & 1.015\% & 1.056\% & 0.9814\% & 1.023\% \\
        \textbf{IAE-Net (ResNet)} & 1.919\% & 1.863\% & 1.702\% & 1.681\% \\
        \hline
        \textbf{IAE-Net} & 0.9639\% & 0.962\% & 0.9327\% & 0.9337\% \\
    \end{tabular}
    \label{tab:darcybenchmark}
\end{table}

\subsection{Scattering Problem}

This section shows the results used to create the figures in Section \ref{sec:scattering}.

\begin{table}[H]
    \centering
    \caption{Results on the forward scattering problem. Model is trained with $s=81$ and tested on different resolutions. Errors are the $L2$ relative error described in Equation \eqref{eqn:l2loss}, scaled by $1e^2$ to show percentage error.}
    \begin{tabular}{c|ccccc}
        \textbf{Model Name} & \textbf{27} & \textbf{41} & \textbf{81} & \textbf{161} & \textbf{241} \\
        \hline
        $FNO$ & 100.1\% & 43.36\% & 16.3\% & 34.22\% & 91.92\% \\
        $FT$ & 197.4\% & 111.9\% & 24.16\% & 171.2\% & 157.3\% \\
        $GT$ & 1182\% & 107.6\% & 22.7\% & 204.3\% & 1131\% \\
        $ResNet+Interpolation$ & 16.4\% & 4.903\% & 4.012\% & 4.22\% & 12.14\% \\
        \hline
        \textbf{IAE-Net (No Skip)} & 1.627\% & 1.71\% & 1.934\% & 2.016\% & 3.131\% \\
        \textbf{IAE-Net (ResNet)} & 1.57\% & 1.812\% & 2.016\% & 2.013\% & 3.98\% \\
        \hline
        \textbf{IAE-Net} & 1.506\% & 1.677\% & 1.82\% & 1.963\% & 2.958\% \\
    \end{tabular}
    \label{tab:scatteringbenchmarkforward}
\end{table}

\begin{table}[H]
    \centering
    \caption{Results on the inverse scattering problem. Model is trained with $s=81$ and tested on different resolutions. Errors are the $L2$ relative error described in Equation \eqref{eqn:l2loss}, scaled by $1e^2$ to show percentage error.}
    \begin{tabular}{c|ccccc}
        \textbf{Model Name} & \textbf{27} & \textbf{41} & \textbf{81} & \textbf{161} & \textbf{241} \\
        \hline
        $FNO$ & 121.4\% & 38.85\% & 14.89\% & 37.48\% & 98.25\% \\
        $FT$ & 182.1\% & 142.1\% & 25.88\% & 136.4\% & 157\% \\
        $GT$ & 2084\% & 248.3\% & 23.61\% & 213.7\% & 1866\% \\
        $ResNet+Interpolation$ & 17.7\% & 3.915\% & 3.936\% & 4.184\% & 11.11\% \\
        \hline
        \textbf{IAE-Net (No Skip)} & 1.456\% & 1.641\% & 1.84\% & 1.927\% & 4.246\% \\
        \textbf{IAE-Net (ResNet)} & 1.574\% & 1.736\% & 1.99\% & 2.123\% & 2.846\% \\
        \hline
        \textbf{IAE-Net} & 1.603\% & 1.666\% & 1.821\% & 1.981\% & 2.77\% \\
    \end{tabular}
    \label{tab:scatteringbenchmarkinverse}
\end{table}

\begin{table}[H]
    \centering
    \caption{Results on the scattering data set using different data generated with different resolutions. Each resolution consists of 10000 data, forming a combined total of 30000 data. Sequential training is done by training the model through loading the data sets in sequence. The last data set in the sequential model is the size $s=162$ data set. Errors are the $L2$ relative error described in Equation \eqref{eqn:l2loss}, scaled by $1e^2$ to show percentage error.}
    \begin{tabular}{c|ccc}
        \textbf{Model Name} & \textbf{81} & \textbf{108} & \textbf{162} \\
        \hline
        \textbf{IAE-Net (Sequential)} & 21.76\% & 14.2\% & 11.12\% \\
        \hline
        \textbf{IAE-Net (No Skip)} & 7.884\% & 8.372\% & 8.338\% \\
        \textbf{IAE-Net (ResNet)} & 5.022\% & 6.064\% & 5.291\% \\
        \hline
        \textbf{IAE-Net} & 4.026\% & 4.773\% & 4.092\% \\
    \end{tabular}
    \label{tab:scatteringmixed}
\end{table}

\subsection{Ellipses}

This section shows the results used to create the figures in Section \ref{sec:ellipses}.

\begin{table}[H]
    \centering
    \caption{Results on the ellipses data set. Model is trained with $s=128$ and tested on different resolutions. Errors are the $L2$ relative error described in Equation \eqref{eqn:l2loss}, scaled by $1e^2$ to show percentage error.}
    \begin{tabular}{c|cccc}
        \textbf{Model Name} & \textbf{s=32} & \textbf{s=64} & \textbf{s=128} & \textbf{s=256} \\
        \hline
        $FNO$ & 26.37\% & 19.77\% & 17.82\% & 20.13\% \\
        $FT$ & 26.89\% & 18.17\% & 14.82\% & 25.26\% \\
        $GT$ & 27.7\% & 22.07\% & 22.19\% & 26.33\% \\
        $ResNet+Interpolation$ & 8.024\% & 7.419\% & 7.16\% & 8.112\% \\
        \hline
        \textbf{IAE-Net} & 6.971\% & 6.289\% & 5.993\% & 6.106\% \\
    \end{tabular}
    \label{tab:ellipsesbenchmark}
\end{table}

\subsection{Fecgsyndb}

This section shows the results used to create the figures in Section \ref{sec:fecgsyndb}.

\begin{table}[H]
    \centering
    \caption{Results on the fecgsyndb data set for proposed model using different resolutions. The model is trained with $s=2000$ and tested on different resolutions. Errors are the $L2$ relative error described in Equation \eqref{eqn:l2loss}, scaled by $1e^2$ to show percentage error.}
    \begin{tabular}{c|cccccc}
        \textbf{Model Name} & \textbf{250} & \textbf{500} & \textbf{1000} & \textbf{2000} & \textbf{4000} \\
        \hline
        $FNO$ & 45.07\% & 24.75\% & 16.76\% & 15.97\% & 18.23\% \\
        $GT$ & 45.30\% & 27.24\% & 18.97\% & 17.75\% & 19.2\% \\
        $DeepONet^{\dag}$ & & & & 99.99\% & \\
        $Unet$ & 112\% & 101\% & 68.78\% & 8.274\% & 69.85\% \\
        $ResNet+Interpolation$ & 66.37\% & 43.73\% & 32.13\% & 31.16\% & 31.92\% \\
        \hline
        \textbf{IAE-Net (No Skip)} & 15.36\% & 10.68\% & 8.723\% & 7.904\% & 8.153\% \\
        \textbf{IAE-Net (ResNet)} & 14.08\% & 9.924\% & 7.925\% & 7.15\% & 7.192\% \\
        \hline
        \textbf{IAE-Net} & 12.08\% & 8.638\% & 7.048\% & 6.802\% & 6.848\% \\
    \end{tabular}
    \label{tab:fecgsyndbdiscretizationinvariance}
\end{table}

\end{document}